\documentclass[lettersize,journal]{IEEEtran}
\usepackage{amsmath,amsfonts}
\usepackage{algorithm}
\usepackage{algpseudocode}
\usepackage{array}
\usepackage{textcomp}
\usepackage{stfloats}
\usepackage{url}
\usepackage{verbatim}
\usepackage{graphicx}
\usepackage{cite}
\usepackage{hyperref}
\usepackage{xcolor}
\usepackage{harpoon}
\usepackage{mathtools}
\usepackage{multirow}
\usepackage{enumitem}
\usepackage{float}

\usepackage[font=small]{caption}
\usepackage[font=scriptsize,labelformat=simple]{subcaption}

\makeatletter
\DeclareRobustCommand{\hvec}[1]{{\mathpalette\hvec@{#1}}}
\newcommand{\hvec@}[2]{%
  \vbox{\offinterlineskip
    \ialign{%
      \hfil##\hfil\cr
      $\m@th#1{}_{\rightharpoonup}$\kern-\scriptspace\cr
      $\m@th#1#2$\cr
    }%
  }%
}
\makeatother
\begin{document}

\title{Goal-oriented Semantic Communication for Robot Arm Reconstruction in Digital Twin: Feature and Temporal Selections}

\author{Shutong Chen, 
Emmanouil Spyrakos-Papastavridis,~\IEEEmembership{Member,~IEEE},
Yichao Jin,~\IEEEmembership{Senior Member,~IEEE}, \\
and Yansha Deng,~\IEEEmembership{Senior Member,~IEEE}
\thanks{S. Chen, E. Spyrakos-Papastavridis and Y. Deng are with the Department of Engineering, King’s College London, Strand, London WC2R 2LS, U.K. (e-mail: shutong.chen@kcl.ac.uk; emmanouil.spyrakos@kcl.ac.uk; yansha.deng@kcl.ac.uk) (Corresponding author: Yansha Deng).

Y. Jin is with ZhejiangLab, Hangzhou 311100, China. (e-mail: yichao.jin@zhejianglab.org). This work was done while he was with the Bristol Research and Innovation Laboratory, Toshiba Europe Ltd., Avon, Bristol, BS1 4ND,  U.K.

This work was supported by the Engineering and Physical Sciences Research
Council (EPSRC) Industrial CASE Doctoral Research Studentship and Toshiba
Europe Ltd. This work was also supported by EPSRC, U.K., under Grant
EP/W004348/1.
}}


\maketitle

\begin{abstract}
As one of the most promising technologies in industry, the Digital Twin (DT) facilitates real-time monitoring and predictive analysis for real-world systems by precisely reconstructing virtual replicas of physical entities. However, this reconstruction faces unprecedented challenges due to the ever-increasing communication overhead, especially for digital robot arm reconstruction. To this end, we propose a novel goal-oriented semantic communication (GSC) framework to extract the GSC information for the robot arm reconstruction task in the DT, with the aim of minimising the communication load under the strict and relaxed reconstruction error constraints. Unlike the traditional reconstruction framework that periodically transmits a reconstruction message for real-time DT reconstruction, our framework implements a feature selection (FS) algorithm to extract the semantic information from the reconstruction message, and a deep reinforcement learning-based temporal selection algorithm to selectively transmit the semantic information over time. We validate our proposed GSC framework through both Pybullet simulations and lab experiments based on the Franka Research 3 robot arm. For a range of distinct robotic tasks, simulation results show that our framework can reduce the communication load  by at least 59.5\% under strict reconstruction error constraints and 80\% under relaxed reconstruction error constraints, compared with traditional communication framework. Also, experimental results confirm the effectiveness of our framework, where the communication load is reduced by 53\% in strict constraint case and 74\% in relaxed constraint case. The demo is available at:  \url{https://youtu.be/2OdeHKxcgnk}.
\end{abstract}


\section{Introduction}
As an emerging paradigm in industry, the Digital Twin (DT) is envisioned to enhance operational safety and reliability through integrating the physical world and digital world \cite{8477101}. By reconstructing high-fidelity digital models of physical entities, DTs can comprehensively replicate real-world systems, thus enabling functionalities such as fault diagnosis \cite{10195218}, predictive maintenance \cite{9690056}, and optimal decision making \cite{10522623}.
However, the real-time reconstruction of digital models normally requires intensive communication resource, posing significant challenges to existing wireless networks \cite{10335924}. 
This challenge becomes even more severe for digital robot arm reconstruction due to its high operating frequency and complex dynamics.
It has been shown that a single control message can take up more than 100 bytes \cite{9013428}, and the throughput requirement for reconstructing an industrial-grade dual robot arm operated at 1ms scale in the DT can reach 138,500 bytes per second \cite{duan2023digital}, not to mention its extension to large-scale industrial scenarios with thousands of robot arms operating simultaneously.
Meanwhile, observations from existing testbed indicate that the frequent control message transmissions without considering the importance of data can lead to data congestion due to the buffer overflow at the receiver end \cite{9417100}.
Therefore, there is an urgent need to reduce the communication load for robot arm reconstruction in DTs.

To tackle this, existing works mainly focused on designing the goal-oriented/task-oriented framework for DT reconstruction, with the aim of reducing communication cost while maintaining the reconstruction error in acceptable levels. In the context of robot arm DT, reconstruction error is commonly characterised by effectiveness-level performance metrics such as the mean square error (MSE) \cite{9530501}\cite{9953092} and Euclidean Distance \cite{10370739}. 
The authors of \cite{9530501} implemented movement prediction algorithms to compensate for packet loss and mitigate the Root-MSE (RMSE) between the digital robot arm trajectory and the ground truth. This method was further exploited to develop communication and prediction co-design frameworks, where deep reinforcement learning (DRL) algorithms were used to minimise the communication load subject to the MSE constraint on robot arm trajectory difference \cite{9953092} or the Euclidean distance constraint between the positions and orientations of the physical robot and its DT \cite{10370739}. Similar to our work, the above goal-oriented frameworks aim to select and transmit data that could steer the system toward achieving the communication goals. However, there still exist two research gaps, which are bridged by our proposed framework in this work.
First, the trajectory difference cannot comprehensively describe the DT reconstruction quality since the robot dynamics (e.g., velocity difference) might introduce deviations in the reconstruction process, this motivates us to refine the effectiveness-level design in this work. Second, these frameworks tend to transmit all generated messages without considering the significance and usefulness of their contents, i.e., the semantic-level problem is unsolved.

To tackle the potential network congestion and increased latency caused by excessive message transmission in goal-oriented frameworks \cite{SeCom}, the concept of semantic communication has been proposed\cite{10251844}. By extracting and transmitting the semantic information behind the original bits, it can significantly reduce the length or transmission frequency of original messages.
Prior works have applied semantic communication to a wide range of traditional data types, such as text \cite{10486856}, speech \cite{9450827}, image \cite{9953076}, and video \cite{9955991}. Meanwhile, semantic communication has also been leveraged to filter and compress the control message or sensor data in reconstruction tasks \cite{10371380,10437950, 10234392,10164147,9380190}. The Age of Information (AoI) \cite{10371380} and its variants (e.g., Age of Incorrect Information \cite{10437950} and ultra-low AoI \cite{10234392}) are the most common semantic metrics for these time-critical data types, which in essence characterise the data timeliness. In \cite{10164147}, the authors defined the semantic value of control messages as the AoI and the similarity between adjacent data. The authors in \cite{9380190} were among the first to study semantic communication for DT robot arm reconstruction, where semantic value is decided based on the usefulness of the message content for robot control. However, they only considered wired connection and their approach might cause the DT to lag behind the real-world. The aforementioned works \cite{10486856, 9450827,9953076,9955991,10371380,10437950, 10234392,10164147,9380190} have explored the benefits of semantic communication extensively, but inevitably share a common limitation that motivates our proposed method in this work. That is, they ignored that the semantic value of messages is not only dependent on their context, but also closely coupled with the specific communication goal.


To take the advantages of both the goal-oriented framework and semantic communication, an integrated goal-oriented semantic communication (GSC) framework was proposed in \cite{zhou2022task}. It jointly considers the semantic-level information and effectiveness-level performance metrics for multi-modal data in various tasks. 
This framework was further implemented in the context of UAV trajectory control \cite{10618994}. The authors utilised a joint function of AoI and Value of information to identify the most important control and command data, with the GSC goal of minimising the trajectory MSE.
Another GSC framework extending from \cite{zhou2022task} was proposed for the point cloud-based avatar reconstruction in the Metaverse \cite{10577270}, where only the critical nodes of the avatar skeleton graph are transmitted to minimise bandwidth usage.
It can be seen that the GSC framework has been developed for various scenarios, but its application for robot arm reconstruction in DT has never been studied yet. Also, existing effectiveness-level performance metrics (e.g., positioning error) for robotic systems cannot fully describe the communication goals of robot arm reconstruction, and the timeliness-centric semantic-level information for control messages cannot fully capture message importance.

Motivated by this, we propose a novel GSC framework with new effectiveness-level and semantic-level designs for robot arm reconstruction in DT. Compared with the existing frameworks \cite{9530501,9953092,10370739, 9380190}, we not only analyse the specific contents of reconstruction messages to identify the most critical GSC information for transmission, but also perform temporal selection to transmit only the most important messages at critical moments without degrading the reconstruction quality. Additionally, we incorporate the velocity difference between the physical and digital robots into the effectiveness-level metrics, to ensure the DT and the physical world share the same dynamics during the reconstruction process.
Our main contributions are summarised as follows:
\begin{itemize} 
    \item We consider a real-time DT reconstruction task for a physical robot arm that performs three different  tasks, including pick-and-place, pick-and-toss and push-and-pull tasks. The goal is to minimise the communication load under the DT reconstruction error constraints.
    \item We jointly consider the impact of robot dynamics and kinematics, and define the effectiveness-level performance metrics as the reconstruction error, which includes the joint angle error and joint velocity error of the robot arm.  At the semantic-level, we reveal that the significance of different message contents (i.e., different features\footnote{We refer to features as different components of the reconstruction message in this work, not the general concept of features in machine learning.}) changes based on robot's current movement. We also capture the temporal features of the reconstruction messages and show that dropping less useful messages barely affect real-time DT reconstruction.
    \item We propose a GSC reconstruction framework that incorporates the effectiveness-level and semantic-level designs to solve the reconstruction problem. Specifically, we develop a Feature Selection (FS) algorithm that can divide the robotic task into several phases (e.g., \textit{grasp} and \textit{release}), and only transmit features that contain GSC information for the current phase. Building upon this, a Proportional-Integral-Derivative-based Primal-dual Deep Q-Network (PPDQN) algorithm is designed to discard the redundant or less useful messages in certain time slots.
    \item Our proposed framework is validated via both Pybullet simulations and experiments using the Franka Research 3 robot arm. The results show that our framework can effectively reduce the communication load while achieving comparable reconstruction accuracy compared to the baseline framework.
\end{itemize}

The rest of this paper is organised as follows: Section \ref{sec:2} presents the system model and the formulated problem. Section \ref{sec:3} and Section \ref{sec:4} introduce the traditional DT reconstruction framework and our proposed GSC reconstruction framework. In Section \ref{sec:sim}, the results of the simulations and physical experiments are presented to verify our proposed framework. Finally, we conclude our work in Section \ref{sec:6}.

\section{System Model and Problem Formulation} \label{sec:2}
In this section, we first present the system model for robot arm reconstruction in DT, and then the wireless channel, and last the reconstruction problem formulation.

\subsection{General Reconstruction System Model}
\begin{figure}
\centering
\setlength{\belowcaptionskip}{-0.5cm} 
\includegraphics[width=0.8\linewidth]{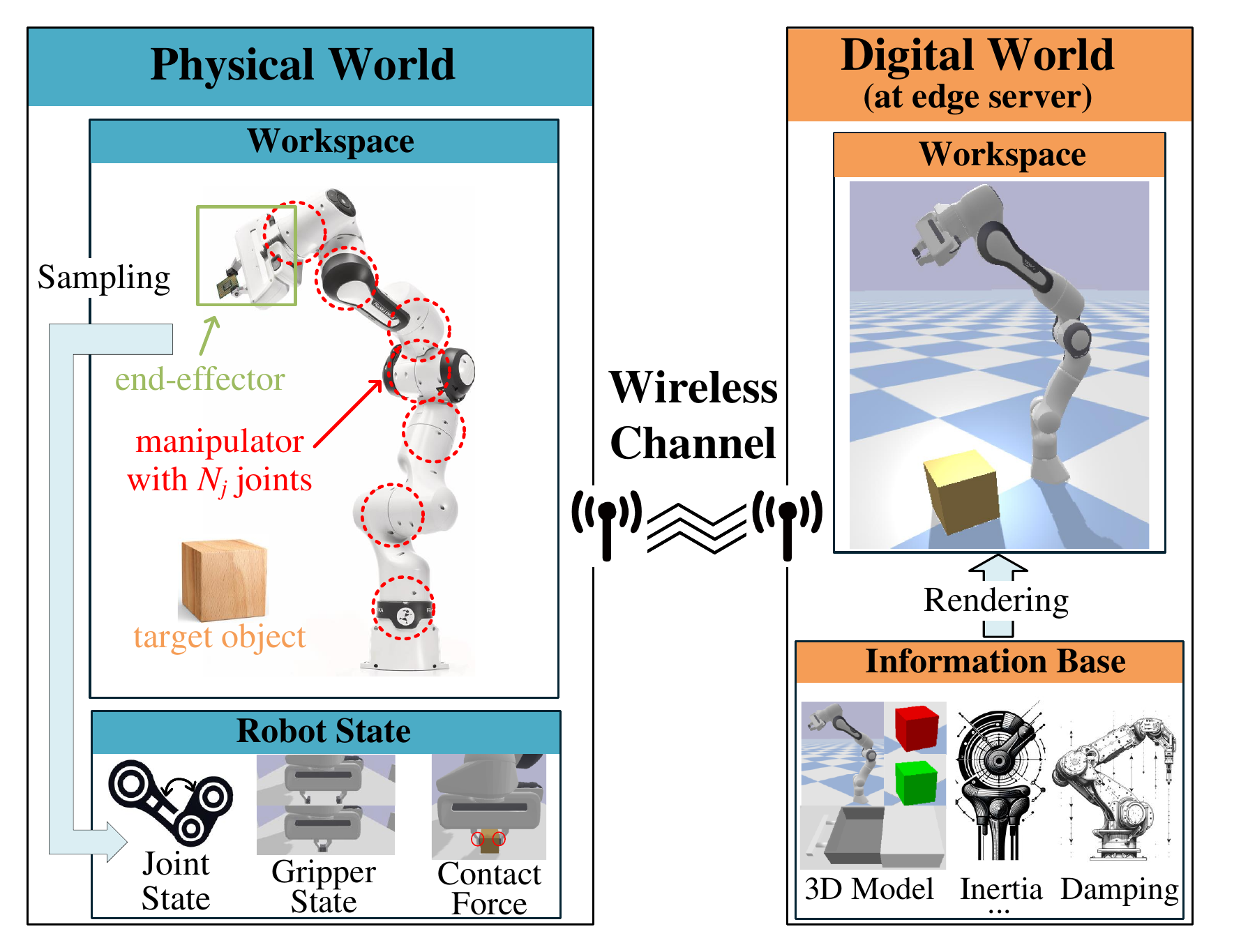}
\caption{Traditional framework for robot arm reconstruction.}
\label{fig:system_trad}
\end{figure}

As illustrated in Fig. \ref{fig:system_trad}, we consider the uplink transmission of the DT system, where the physical world is wirelessly connected with the digital world deployed at the edge server. Specifically, the physical world contains a robot arm and several target objects, whereas the digital world stores DT models, with each model paired with a corresponding physical entity. 
The physical robot arm periodically sends the digital world messages that contain data needed for reconstruction, aiming to reconstruct its DT and its interaction environment in the digital world. 

In the physical world, the robot arm, that is equipped with an $N_j$ degree-of-freedom (DoF) manipulator and also a two-finger gripper as its end-effector, performs robotic tasks (e.g., component assembly and product sorting) involving target objects. 
The digital world is assumed to cache an information base that stores the 3-dimensional (3D) digital models and physical properties (e.g., mass and stiffness) of all relevant entities in the physical world, where these knowledge can be shared before the start of transmission. In this way, the initial 3D models of all the relevant physical world entities are rendered in the digital world.
During the reconstruction process, since the digital robot arm synchronises its motion to mirror that of the physical arm, the DTs of the target objects also need to be updated during their interaction with the digital robot arm.



\subsection{Channel Model}
To practically model the dynamic real-world transmission environment, we model the wireless channel between the physical world and digital world as Nakagami-$m$ fading. 
It is particularly suitable for industrial wireless settings as it can characterize a wide range of fading models (e.g., Rayleigh and Rician fading), this enables more robust simulations that can model  the dynamic small-scale fading caused by interference, obstacles, and movement within the industrial environment.
The channel fading power gain $g$ follows the Nakagami-$m$ distribution and its probability density function is \cite{7045615}
\begin{align} 
	f_G \left ({g}\right) = \frac{g^{m-1}}{\Gamma(m)} \left(\frac{m}{\Omega}\right)^{m} e^{-{m\over\Omega}g},
\end{align}
where $\Gamma(m)$ refers to the Gamma function, while $m$ and $\Omega$ are the shape parameter and the scale parameter, respectively.

Meanwhile, all the packets are assumed to experience large-scale path loss with coefficient $\alpha$, and the overall channel power gain can be expressed as 
\begin{align}
	h =  d^{-\alpha} \mathbb{E} \left[\vert g \vert^2 \right], \label{equ:fading}
\end{align}
where $d$ is the distance between the physical robot and the edge server. 
Accordingly, the system signal-to-noise ratio (SNR) is derived as $\text{SNR} = {Ph}/{\sigma^2}$, where $P$ is the transmit power and $\sigma^2$ denotes the Gaussian white noise power. 

We also assume that the DT can decode the received packet only if the SNR is above a threshold $\beta$. 
This is used to simulate the sensing ability of the real-world receiver. When the SNR falls below the threshold, the signal is corrupted by noise and becomes difficult to decode.
Thus, the effect of the wireless channel is denoted by a binary variable 
$\delta^c$ using
\begin{align}
\setlength{\abovedisplayskip}{-3cm}
\setlength{\belowdisplayskip}{-3cm}
	\delta^c=\left\{
	\begin{array}{ll}
		1, \quad \quad  \text{SNR} \geq \beta, \\
		0, \quad \quad \text{else},\\
	\end{array} \right.
\end{align}
in which $\delta^c = 1$ indicates successful packet delivery, while $\delta^c = 0$ indicates failure packet delivery. 

Note that the edge server with the digital world will send an acknowledgement (ACK) packet back once the  message is successfully received. We also assume that the downlink transmission from the edge server to the physical robot is ideal \cite{9170549}, so as to focus on the uplink transmission for DT reconstruction instead. In practice, this assumption can be supported by many techniques, such as Hybrid Automatic Repeat Request (HARQ) with incremental redundancy \cite{HARQ} and proactive transmission \cite{retrans}, which are specifically designed to support the ultra-reliable transmission of short packets.

\vspace{-0.1cm}
\subsection{Problem Formulation}

We aim to reconstruct a DT of the real-world robot arm that can accurately replicate its dynamics, states, and real-time motions with reduced communication cost.
To this end, the main objective is to minimise the communication load subject to the DT reconstruction error constraints. 

We assume that time is discretised into $N_T$ independent slots, where each slot is indexed by $t$, for $t \in [0, N_T]$ and $N_T \in \mathbb{Z}$. The communication load $L_t$ is defined as the number of bits transmitted during the $t$-th slot. We also define the normalised joint angle error using $e_{q_t}$ and the normalised joint velocity error using $e_{\dot{q}_t}$ to evaluate the DT reconstruction error within each time slot based on
\begin{align}
    e_{q_t} = \frac{1}{N_j}\sum_{k=1}^{N_j}\frac{q_t^k-\hat{q}_t^k}{q_{\text{max}}^k - q_{\text{min}}^k}, \quad
    e_{\dot{q}_t} = \frac{1}{N_j}\sum_{k=1}^{N_j}\frac{\dot{q}_t^k-\hat{\dot{q}}_t^k}{\dot{q}_{\text{max}}^k - \dot{q}_{\text{min}}^k},  \label{equ:error}
\end{align}
where $q_t^k$ and $\dot{q}_t^k$ are the physical robot's joint angle and joint velocity for the $k$-th joint, $\hat{q}_t^k$ and $\hat{\dot{q}}_t^k$ are the corresponding values for the digital robot.
$q_{\text{max}}^k$, $q_{\text{min}}^k$, $\dot{q}_{\text{max}}^k$ and $\dot{q}_{\text{min}}^k$  correspond to the maximum and minimum angle and velocity limits of the $k$-th joint, respectively. 

As the aim is to minimise the average communication load $L_i$ over all time slots subject to the DT reconstruction error, the problem is mathematically formulated as
\begin{align}
    \min\limits  & \lim_{N_T \to \infty} \frac{1}{N_T} \sum^{N_T}_{i=1} L_t, \label{equ:problem} \\ 
    \textrm{s.t.} & \quad  e_{q_t} \leq  C_{q_t}, \quad \forall i \in N_T,              \nonumber \\
    & \quad  e_{\dot{q}_t} \leq  C_{\dot{q}_t}, \quad \forall i \in N_T,              \nonumber
\end{align}  
where $C_{q_t}$ and $C_{\dot{q}_t}$ are the error constraints on the joint angle and joint velocity, respectively.

\begin{figure*}[h]
\centering
\setlength{\belowcaptionskip}{-0.3cm} 
\includegraphics[width=0.9\linewidth]{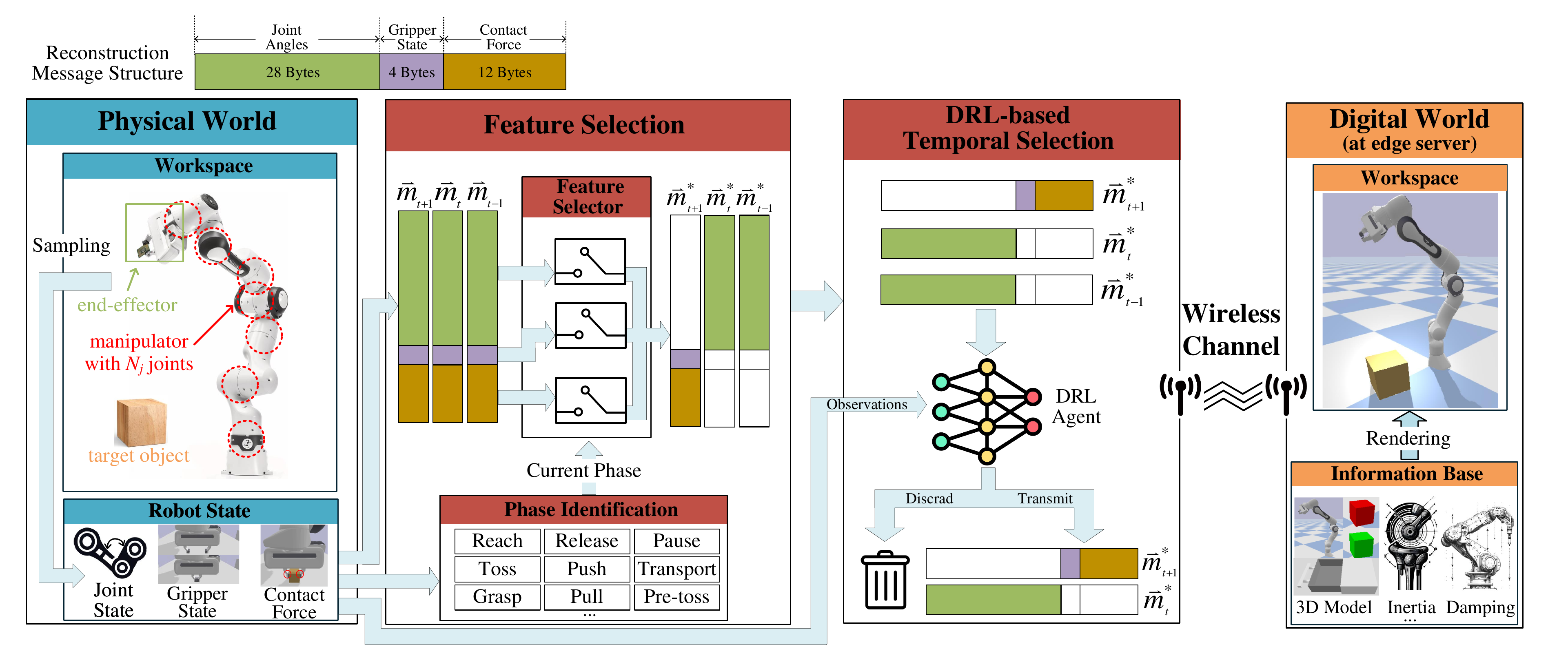}
\caption{Our proposed GSC framework for robot arm reconstruction.}
\label{fig:system_GSC}
\end{figure*}
\section{Traditional DT Reconstruction Framework} \label{sec:3}
Conventionally, the physical robot arm's operational state is recorded in the message $\hvec{m}$ at the beginning of every time slot, which is then transmitted to the digital world. Based on this, the digital robot arm model is reconstructed, and its latest motion is rendered. The message $\hvec{m}_t$ transmitted by the robot arm in the physical world at the $t$-th time slot is defined as
\begin{align}
\label{equ:msg}
    \hvec{m}_t = (\mathcal{Q}_{t}, \delta^g_t, \hvec{f_{t}}), 
\end{align}
where features $\mathcal{Q}_{t}, \delta^g_t, \hvec{f_{t}}$, and other important variables are defined as follows: 
\begin{itemize}
\item We define $\mathcal{Q}_{t} = \{q_{t}^1, q_{t}^2, ..., q_{t}^{N_j}\}$ as the joint angle set that describes the positions of all $N_j$ joints. We also denote the joint angular velocity set as $\dot{\mathcal{Q}}_{t} = \{\dot{q}_{t}^1, \dot{q}_{t}^2, ..., \dot{q}_{t}^{N_j}\}$ to characterise the robot arm's motion.
\item We define the binary gripper state of the physical robot as $\delta^g_t \in \{0, 1\}$, where $0$ denotes an open gripper and $1$ indicates a closed one. Also, the  gripper state of the digital robot arm is defined as  $\hat{\delta}^g_{t-1}$.
\item We denote the three-axis contact force vector between the gripper and the target object using $\hvec{f_t} = (f_x, f_y, f_z )$, which can be measured by an external force sensor. The resolution (i.e., minimum detectable force) and force derivative threshold (i.e., maximum detectable noise) of the force sensor are defined as $\rho_1$ and $\rho_2$ \footnote{The resolution $\rho_1$ is normally specified in the technical datasheet of the force sensor. The force derivative threshold $\rho_2$ is set to twice the maximum noise amplitude of the sensor, where the maximum noise amplitude is defined according to the sensor accuracy.}
, respectively.
\end{itemize}

We adopt the standard message format specified in the Robot Operating System (ROS) \cite{quigley2009ros}, therefore the sizes of the joint angle set, 3-axis contact force vector, and gripper state are set to 28 bytes, 12 bytes, and 4 bytes, respectively. Consequently, the total size of the original reconstruction message is 44 bytes.

However, it is noteworthy that this traditional framework can potentially lead to unnecessary data acquisition and transmission in both the feature domain and time domain.
In the feature domain, the traditional framework transmits the complete reconstruction messages, including contents that are not necessarily useful to the reconstruction. For instance, when the physical robotic arm is approaching the object, its gripper state remains unchanged and is in fact irrelevant to the reconstruction accuracy.
In the time domain, the traditional framework transmits reconstruction messages at every time slot, without considering the fact that some messages could be discarded without affecting the reconstruction accuracy. This is because the DT shares the similar physical rules as the real-world system, which allows the digital robot to maintain its desired trajectory and speed without constant updates.
Therefore, it is clear that the traditional reconstruction framework would result in communication resource wastage due to the excessive transmission of uninformative messages. To address this, we propose a GSC reconstruction framework to minimise the communication load under the reconstruction error constraints.

\section{The Proposed GSC Reconstruction Framework} \label{sec:4}
In this section, we provide a detailed description of our proposed GSC reconstruction framework, as illustrated in Fig. \ref{fig:system_GSC}. Compared with the traditional framework, we first develop a FS algorithm in the feature domain to process each message $\hvec{m}_t$ and filter out irrelevant features that do not contribute to the reconstruction accuracy. Subsequently,  a PPDQN algorithm is designed in the time domain to decide whether it is necessary to trigger the transmission in the current time slot. 

\subsection{Feature Selection} 
The traditional framework treats every feature equally without considering their importance to the task, which results in transmission resource wastage in unnecessary and useless features. 
In effect, it is noted that the importance of different features contained in message $\hvec{m}_t$ changes over time based on the robot's motions. Motivated by this, we design a FS algorithm that can segment these motions into different phases. By analysing the current dynamic state (e.g., joint velocities) of the physical robot, this algorithm can identify the current phase and selectively transmit features that are most useful for the reconstruction in that phase. Consequently, the content of the output semantic message $\hvec{m}_t^*$ is dynamically adjusted at every time slot, but always contains only GSC information.


Specially, we develop the FS algorithm for three tasks commonly considered by the robotics society, which are pick-and-place, pick-and-toss, and push-and-pull.
They are the most fundamental, and commonly executed robot arm tasks in both industrial scenarios and home environments \cite{toss}\cite{tidybot}.

\subsubsection{Pick-and-place}
The pick-and-place task involves manipulating objects through six phases: \textit{reach} phase, \textit{grasp} phase, \textit{transport} phase, \textit{pre-release} phase, \textit{release} phase and \textit{pause} phase, as illustrated in Fig. 3(a). 
To identify these phases, the FS algorithm first evaluates the following variables that will be utilised as identification conditions: the Cartesian end-effector velocity $\hvec{v}^{\text{ee}}_t = ({v}^{x}_{t}, {v}^{y}_t, {v}^{z}_t)$ and its norm; the norm of the contact force vector $\|\hvec{f_{t}}\|$ and its derivative $\|\hvec{f_{t}}\|^{\prime}$; and the gripper width $r_t$ as well as its derivative $r^{\prime}_t$. Specifically, the end-effector velocity cannot be obtained directly without external motion capture devices, but can be derived through the robot forward kinematics equation \cite{kinematics}
\begin{align} \label{equ:ee}
\hvec{v}^{\text{ee}}_t = \boldsymbol{J}(\mathcal {Q}_t) \dot{\mathcal{Q}}_{t},
\end{align}
where $\boldsymbol {J}(\mathcal {Q}_t)$ denotes the Jacobian matrix\footnote{The Jacobian matrix is a fundamental concept in robotics that relates the end-effector velocity to the joint velocities; herein, the detailed expression is omitted due to the page limit.} of the robot arm.

\begin{figure*}
\centering
\setlength{\belowcaptionskip}{-0.3cm} 
\includegraphics[width=0.9\linewidth]{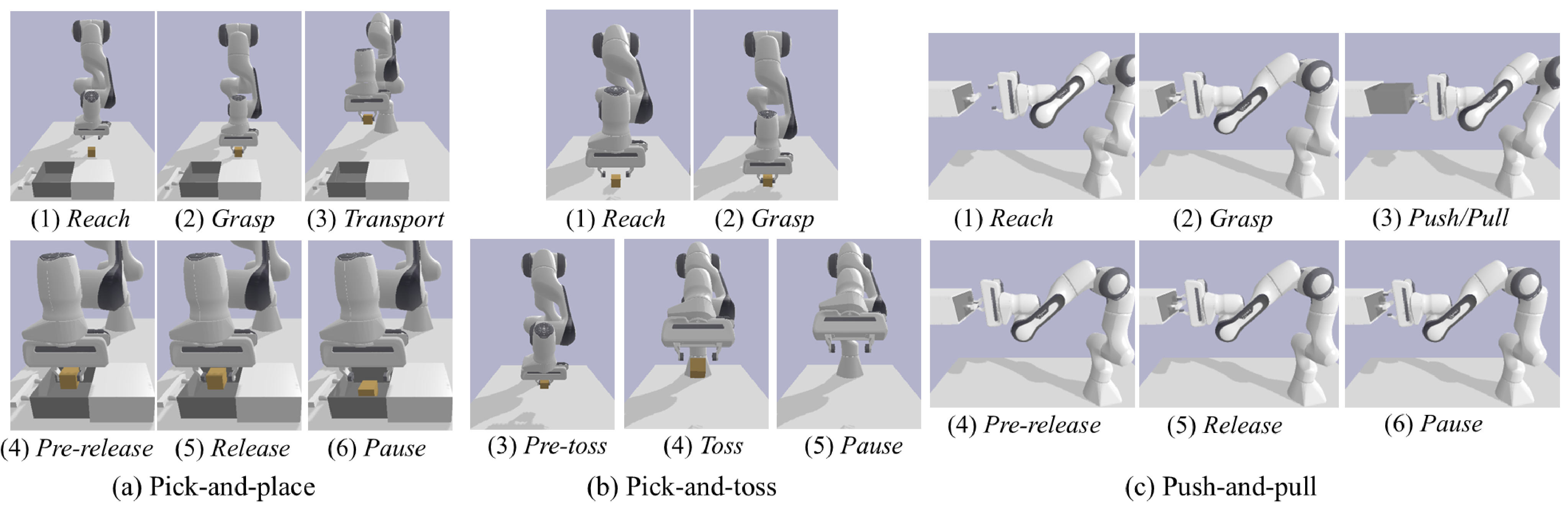}
\caption{Phases of different robot tasks.}
\label{fig:phases}
\end{figure*}

As presented in Algorithm \ref{pickplace}, instead of analysing the algorithm according to the phase execution order, we group phases that send similar features and analyse them together. Specifically, we first consider phases involving manipulator motions, where joint angles must be transmitted for manipulator motion recovery. We then consider the remaining phases, where the gripper is moving and its state must be updated.
\textcolor{black}{
\begin{algorithm}[h]
	\caption{Feature Selection for Pick-and-place Task}\label{pickplace}
	\textbf{Input:} Original message $\hvec{m}_t$ and joint velocity set $\dot{\mathcal{Q}}_{t}$\\
	\textbf{Output:} Semantic message $\hvec{m}_{t}^*$
	\begin{algorithmic}
		\For{$i = 1$ to $N_T$} 
            \State Calculate $\| \hvec{v}^{\text{ee}}_t \|$ using Eq. (\ref{equ:ee}), $\|\hvec{f_{t}}\|$, $\|\hvec{f_{t}}\|^{\prime}$ and $r^{\prime}_t$
            \If{$\|\hvec{v}^{\text{ee}}_t\| > v_\text{th}$}
		\If{$\| \hvec{f_{t}}\| \leq \rho_1 $ } 
		\State \emph{Reach} phase: $\hvec{m}_{t}^* = (\mathcal{Q}_{t})$
		\ElsIf{$\| \hvec{f_{t}}\| > \rho_1 $ \textbf{and} $| \|\hvec{f_{t}}\|^{\prime}| \leq \rho_2$}
		\State \emph{Transport} phase: $\hvec{m}_{t}^* = (\mathcal{Q}_{t}, \hvec{f_{t}})$
            \Else
            \State Raise error: collision detected or object slipped
		\EndIf
        \ElsIf {$\|\hvec{v}^{\text{ee}}_t\| \leq v_\text{th}$ \textbf{and} $\delta^g_t \neq \hat{\delta}^g_{t-1}$}
		\If{$\|\hvec{f_{t}}\|^{\prime} > \rho_2$}
		\State \emph{Grasp} phase: $\hvec{m}_{t}^* = (\delta^g_t, \hvec{f_{t}})$
		\ElsIf{$\| \hvec{f_{t}}\| > \rho_1 $}
		\State \emph{Pre-release} phase: $\hvec{m}_{t}^* = (\delta^g_t)$
  		\ElsIf{$\| \hvec{f_{t}}\| \leq \rho_1 $ \textbf{and}
        $r^{\prime}_t > 0$}
		\State \emph{Release} phase: $\hvec{m}_{t}^* = (\delta^g_t)$
            \Else
            \State Raise error: grasp failure or mishandling
            \EndIf
            \Else
            \State \emph{Pause} phase or Drop gripper state update $\hvec{m}_{t}^* = \o$
		\EndIf
		\EndFor
	\end{algorithmic}
\end{algorithm}}

If the end-effector velocity $\|\hvec{v}^{\text{ee}}_t\|$ exceeds the velocity threshold for transitioning from a static state to motion $v_\text{th}$, the manipulator is currently in motion and the robot is in either the \textit{reach} or \textit{transport} phase. Please note that $v_\text{th}$ is derived based on the positional precision of the robot end-effector, where the level of positional precision determines the magnitude of residual motions when the robot is in the static state.

\textbf{Reach:} The first stage of the pick-and-place task is to reach the target object. Given that the gripper does not make any contact when the robot approaches the object, the current phase is identified as \textit{reach} if the norm of the contact force resides in the vicinity of 0, and remains below the force sensor's resolution $\| \hvec{f_{t}}\| \leq \rho_1 $. During the \textit{reach} phase, the system's focus is  on the motion of the manipulator, and only goal-oriented features, i.e., joint angles $\mathcal{Q}_{t}$, is transmitted for motion recovery, hence the output semantic message $\hvec{m}_{t}^* = (\mathcal{Q}_{t})$. 

\textbf{Transport:} The goal of the \textit{transport} phase is to transport the object to the target location, therefore the focus is still on the manipulator's motion. However, unlike the \textit{reach} phase, there is a measurable, constant contact force being applied to the gripper when the robot moves together with the object. Thus, the current phase is determined as \textit{transport} if a contact is sensed such that $\| \hvec{f_{t}}\| > \rho_1 $, and the derivative of the contact force falls within the force derivative threshold $| \|\hvec{f_{t}}\|^{\prime}| \leq \rho_2$. 

Notably, the contact force $\hvec{f_{t}}$ is also transmitted to the DT alongside the joint angles $\mathcal{Q}_{t}$ during this phase to ensure the construction accuracy, and the contact force has never been considered for robot arm modelling in previous works \cite{9530501, 9953092, 10370739}. In robotic control, the actuators are expected to generate the required joint forces or torques\footnote{Torque is the rotational effect produced by a force, which is equal to the vector cross product of the force and the lever arm (i.e., the distance from the pivot point to where the force is applied).} that would enable the joints to reach the desired angles and track the desired velocities. Specifically, the robot ideally needs to compensate for inertial, Coriolis/centripetal, and gravitational effects, as well as external forces applied to its end-effector or elsewhere on its structure. The relationship between the robot input torques $\hvec{\tau}$ and its joint dynamics is given by\cite{kinematics}
\newcommand{\harpoon}{\overset{\rightharpoonup}}
\begin{align}
    \hvec{\tau} = \underbrace{\boldsymbol{H}(\mathcal {Q})\ddot{\mathcal {Q}}}_{\text{inertial torques}} + \!\!\!\!\!\!\!\!\!\! \overbrace{\boldsymbol{C}(\mathcal {Q},\dot{\mathcal {Q}})\dot{\mathcal {Q}}}^{\text{Coriolis and centripetal torques}} \!\!\!\!\!\!\!\!\!\! + \!\!\!\!\!\!\!\!\!\!\ \underbrace{\hvec{\tau_g}(\mathcal {Q})}_{\text{gravitational torque}} \!\!\!\!\!\!\!\!\!\! + \overbrace{\boldsymbol{J}(\mathcal {Q})^{\mathrm{T}} \hvec{f}_\text{\!\!ext}}^{\text{external torque}},
    \label{equ:tau}
\end{align}
where all four terms are dependent on the joint angles, while the last term is also directly determined by the external force applied to the end-effector $\hvec{f}_\text{\!\!ext}$. When the robot is not in contact, the effect of the external force becomes irrelevant, therefore only transmitting the joint angle is sufficient for accurate DT reconstruction. However, during the \textit{transport} phase when the contact continues to occur, the contact force becomes the main component of the external force, and should thus be transmitted for the DT to calculate the additional required torque. To this end, the output semantic message contains both joint angle set and contact force $\hvec{m}_{t}^* = (\mathcal{Q}_{t}, \hvec{f_{t}})$.


When the end-effector velocity decreases below the moving velocity threshold $\|\hvec{v}^{\text{ee}}_t\| \leq v_\text{th}$, the manipulator is considered to be static, and the movement of the gripper and the updating of gripper state $\delta^g_t$ then become more important. Specifically, we adopt the classical semantics-aware transmission policy \cite{9921185}, which means
updates are only triggered whenever a discrepancy arises between the gripper states of the physical and digital robot arms $\delta^g_t \neq \hat{\delta}^g_{t-1}$. That is, updates occur only when they can reduce the gripper state mismatch. For instance, an update is allowed to be transmitted when the physical robot decides to close the gripper and receives the ACK packet from the DT regarding a previously delivered state message, in which the gripper was still open. In this case, the algorithm prioritises messages that reduce the mismatch between source and receiver, and drops further message transmissions if the previous one is successfully received and the mismatch is already reduced. 

\textbf{Grasp:} During the \textit{grasp} phase, to capture the precise moment when the physical robot arm changes the gripper state $\delta^g_t$, the derivative of the contact force is firstly evaluated. Since the contact force gradually increases with time during the \textit{grasp} phase, the algorithm then uses the contact force derivative's exceedance of the force derivative threshold $\|\hvec{f_{t}}\|^{\prime} > \rho_2$ as a condition, and then transmits the gripper state $\delta^g_t$ as well as the contact force $ \hvec{f_{t}}$ accordingly, with the output semantic message expressed as $\hvec{m}_{t}^* = (\delta^g_t, \hvec{f_{t}})$. 

\textbf{Pre-release and Release:} However, the above force derivative-based method is not applicable to the \textit{release} phase. This is because, for rigid objects, the force change occurs too swiftly to be accurately captured; while for non-rigid objects, the force magnitude fluctuation is normally irregular and noisy. To address this, we divide the release process into two parts according to the contact status. We first define a \textit{pre-release} phase to describe the state between \textit{Transport} and \textit{release} when the gripper is still holding the object $\| \hvec{f_{t}}\| > \rho_1$, but the end-effector velocity is already under the velocity threshold $\|\hvec{v}^{\text{ee}}_t\| \leq v_\text{th}$. Then, the robot executes the actual \textit{release} action, during which the gripper and the object are no longer in contact $\| \hvec{f_{t}}\| \leq \rho_1 $, and the gripper width is increasing $r^{\prime}_t > 0$.
Thus, the noisy release process can be recognised by the algorithm, while the gripper state $\delta^g_t$ continues to be updated until the edge server successfully receives it. The output semantic message is expressed as $\hvec{m}_{t}^* = (\delta^g_t)$.

\textbf{Pause:} When the robot is not in motion and not in contact, the current phase is identified as \textit{pause}, wherein the physical robot remains static and silent. The output semantic message is then an empty set, i.e., $\hvec{m}_{t}^* = \o$.

\subsubsection{Pick-and-toss}
The pick-and-toss task can also be segmented into five phases, i.e.,  \textit{reach} phase, \textit{grasp} phase, \textit{pre-toss} phase, \textit{toss} phase, and \textit{pause} phase, as shown in Fig. 3(b). The phase identification conditions of the \textit{reach} phase, \textit{grasp} phase and \textit{pause} phase are the same as those in the pick-and-place task. However, herein we introduce the \textit{pre-toss} phase and \textit{toss} phase to describe the robot's motion of throwing the target object, as shown in \textbf{Algorithm \ref{alg:toss}}.

\textbf{Pre-toss:} After grasping the object, the robot adjusts its end-effector orientation towards the direction opposite to the tossing action to prepare for the \emph{toss} phase, which also aims to ensure that the object obtains the desired initial velocity. Note that the velocity of the end-effector remains relatively low, $\|\hvec{v}^{\text{ee}}_t\| \leq v_\text{th}$, during this process in order to reach the desired orientation. Meanwhile, the robot holds the object stably with measurable and constant contact force, i.e., $\| \hvec{f_{t}}\| > \rho_1 $ and $| \|\hvec{f_{t}}\|^{\prime}| \leq \rho_2$. The above three conditions are then used for phase identification. Within the output semantic messages, the contact force is discarded because it has almost no variation during this phase, thus only the joint angles are transmitted to synchronise the manipulator's motion, i.e., $\hvec{m}_{t}^* = (\mathcal{Q}_{t})$. 
\begin{algorithm}[htb] 
	\caption{Feature Selection for Pick-and-toss Task}\label{alg:toss}
	\textbf{Input:} Original message $\hvec{m}_t$ and joint velocity set $\dot{\mathcal{Q}}_{t}$\\
	\textbf{Output:} Semantic message $\hvec{m}_{t}^*$
	\begin{algorithmic}
		\For{$i = 1$ to $N_T$} 
            \State Calculate $\| \hvec{v}^{\text{ee}}_t \|$ using Eq. (\ref{equ:ee}), $\|\hvec{f_{t}}\|$, $\|\hvec{f_{t}}\|^{\prime}$ and $r^{\prime}_t$
		\If{$\|\hvec{v}^{\text{ee}}_t\| > v_\text{th}$} 
                \If{$\| \hvec{f_{t}}\| \leq \rho_1 $}
                \State \emph{Reach} phase: $\hvec{m}_{t}^* = (\mathcal{Q}_{t})$
                \ElsIf{$\| \hvec{f_{t}}\| > \rho_1 $ \textbf{or} $\delta^g_t \neq \hat{\delta}^g_{t-1}$}
		    \State \emph{Toss} phase: $\hvec{m}_{t}^* = (\mathcal{Q}_{t}, \delta^g_t)$
                \EndIf
            \Else
                \If{$\| \hvec{f_{t}}\| > \rho_1 $}
		      \State \emph{Pre-toss} phase: $\hvec{m}_{t}^* = (\mathcal{Q}_{t})$
                \ElsIf{$\|\hvec{f_{t}}\|^{\prime} > \rho_2$  \textbf{and} $\delta^g_t \neq \hat{\delta}^g_{t-1}$}
		    \State \emph{Grasp} phase: $\hvec{m}_{t}^* = (\delta^g_t, \hvec{f_{t}})$
                \Else
                \State \emph{Pause} phase or Drop gripper state update $\hvec{m}_{t}^* = \o$
                \EndIf
		\EndIf
		\EndFor
	\end{algorithmic}
\end{algorithm}

\textbf{Toss:} During the \emph{toss} phase, the robot accelerates forward to provide the object with a large initial velocity, then opens the gripper so the object can reach the target location before hitting the ground. High-speed motion of the robot and measurable contact force will be observed within this phase, which means the identification condition is a large end-effector velocity $\|\hvec{v}^{\text{ee}}_t\| > v_\text{th}$ and a constant contact force $\| \hvec{f_{t}}\| > \rho_1 $. Note that the \emph{toss} phase actually contains the release action, hence the transmission should also be triggered if the physical robot has a different gripper state from the digital robot. This also means the gripper state must be transmitted alongside the joint angle set, otherwise both the motion of the robot arm and the landing point of the object will suffer huge deviations due to the asynchronous object release timing.
Thus, the output semantic message is $\hvec{m}_{t}^* = (\mathcal{Q}_{t}, \delta^g_t)$.

\subsubsection{Push-and-pull}
In the push-and-pull task, the robot arm is required to pull the object to the target location, and push it back to its original position. 
We divide this task into seven phases, i.e., \textit{reach} phase, \textit{grasp} phase, \textit{pull} phase, \textit{push} phase,  \textit{pre-release} phase,  \textit{release} phase and \textit{pause} phase, as shown in Fig. 3(c).  
Their identification conditions are similar to the pick-and-place case, with the only difference being that the \emph{transport} phase is replaced by the \emph{push/pull} phase. The FS design of the push-and-pull task is shown in \textbf{ Algorithm \ref{alg:push}}.

\textbf{Push/Pull:} During the \emph{push/pull} phase, the robot pushes or pulls the object towards the desired direction. Similarly to the \emph{transport} phase, the manipulator is in motion while a continuous, and constant contact force can be detected by the external force sensor. However, the difference is that the vertical velocity of the end-effector is relatively small since the robot moves horizontally. Therefore, the condition for the phase identification also includes that the vertical velocity of the end-effector should reside in the vicinity of zero, i.e., $v^z_i < v_\text{th}$. The output semantic message is the same as that in the \emph{transport} phase, in order to mimic the manipulator's motion and monitor the contact status, i.e., $\hvec{m}_{t}^* = (\mathcal{Q}_{t}, \hvec{f_{t}})$. 
\begin{algorithm}[t]
	\caption{Feature Selection for Push-and-pull Task}\label{alg:push}
	\textbf{Input:} Original message $\hvec{m}_t$ and joint velocity set $\dot{\mathcal{Q}}_{t}$\\
	\textbf{Output:} Semantic message $\hvec{m}_{t}^*$
	\begin{algorithmic}
		\For{$i = 1$ to $N_T$} 
            \State Calculate $\| \hvec{v}^{\text{ee}}_t \|$ using Eq. (\ref{equ:ee}), $\|\hvec{f_{t}}\|$, $\|\hvec{f_{t}}\|^{\prime}$ and $r^{\prime}_t$
            \If{$\|\hvec{v}^{\text{ee}}_t\| > v_\text{th}$}
		\If{$\| \hvec{f_{t}}\| \leq \rho_1 $ } 
		\State \emph{Reach} phase: $\hvec{m}_{t}^* = (\mathcal{Q}_{t})$
		\ElsIf{$\| \hvec{f_{t}}\| > \rho_1 $, $| \|\hvec{f_{t}}\|^{\prime}| \leq \rho_2$ \textbf{and} $v^z_i < v_\text{th}$}
		\State \emph{Push/Pull} phase: $\hvec{m}_{t}^* = (\mathcal{Q}_{t}, \hvec{f_{t}})$
            \Else
            \State Raise error: collision detected or object slipped
		\EndIf
        \ElsIf {$\|\hvec{v}^{\text{ee}}_t\| \leq v_\text{th}$ \textbf{and} $\delta^g_t \neq \hat{\delta}^g_{t-1}$}
		\If{$\|\hvec{f_{t}}\|^{\prime} > \rho_2$}
		\State \emph{Grasp} phase: $\hvec{m}_{t}^* = (\delta^g_t, \hvec{f_{t}})$
		\ElsIf{$\| \hvec{f_{t}}\| > \rho_1 $}
		\State \emph{Pre-release} phase: $\hvec{m}_{t}^* = (\delta^g_t)$
  		\ElsIf{$\| \hvec{f_{t}}\| \leq \rho_1 $ \textbf{and}
        $r^{\prime}_t > 0$}
		\State \emph{Release} phase: $\hvec{m}_{t}^* = (\delta^g_t)$
            \Else
            \State Raise error: grasp failure or mishandling
            \EndIf
            \Else
            \State \emph{Pause} phase or Drop gripper state update $\hvec{m}_{t}^* = \o$
		\EndIf
		\EndFor
	\end{algorithmic}
\end{algorithm} 

\subsubsection{Message Output}
Finally, we summarise the message output for each phase across three tasks in Table \ref{tab:message_output}.
Specifically, during phases involving manipulator motions, joint angles are always transmitted to recover the manipulator's movements, while when the gripper is moving, its state is always updated to reflect its current action. For phases involving interaction with objects, the contact force is additionally transmitted to capture interaction dynamics.
It can be observed that the selection of features in each phase is actually driven by the inherent semantic information (or physical meaning) of these features, which is intrinsically  related to the physical actions of the robot arm during those phases. Compared with existing works that define the GSC information of control data by timeliness or urgency, we emphasise that the usefulness of the information content for the control process is also important.
\begin{table}[t]
\centering
\caption{Message Output for Each Phase in Different Robotic Tasks}
\renewcommand{\arraystretch}{1.4} 
\begin{tabular}{|c|c|c|}
\hline
\textbf{Task}         & \textbf{Phase}        & \textbf{Message Output}            \\ \hline
        & Reach                 & Joint angles $\mathcal{Q}_{t}$               \\ \cline{2-3} 
                    Pick-and-place   & Grasp                 & Gripper state $\delta^g_t$, contact force $\hvec{f_{t}}$ \\ \cline{2-3} 
                   (Algorithm \ref{pickplace})   & Transport             & Joint angles $\mathcal{Q}_{t}$, contact force $\hvec{f_{t}}$ \\ \cline{2-3} 
                      & Release               & Gripper state $\delta^g_t$       \\ \cline{2-3} 
                      & Pause                 & None                               \\ \hline
         & Reach                 & Joint angles $\mathcal{Q}_{t}$               \\ \cline{2-3} 
           Pick-and-toss           & Grasp                 & Gripper state $\delta^g_t$, contact force $\hvec{f_{t}}$ \\ \cline{2-3} 
                   (Algorithm \ref{alg:toss})   & Pre-toss              & Joint angles $\mathcal{Q}_{t}$               \\ \cline{2-3} 
                      & Toss                  & Joint angles $\mathcal{Q}_{t}$, gripper state $\delta^g_t$ \\ \cline{2-3} 
                      & Pause                 & None                               \\ \hline
         & Reach                 & Joint angles $\mathcal{Q}_{t}$               \\ \cline{2-3} 
                Push-and-pull       & Grasp                 & Gripper state $\delta^g_t$, contact force $\hvec{f_{t}}$ \\ \cline{2-3} 
                   (Algorithm \ref{alg:push})   & Push/Pull             & Joint angles $\mathcal{Q}_{t}$, contact force $\hvec{f_{t}}$ \\ \cline{2-3} 
                      & Release               & Gripper state $\delta^g_t$      \\ \cline{2-3} 
                      & Pause                 & None                               \\ \hline
\end{tabular}
\label{tab:message_output}
\end{table}

\subsection{PID-based primal-dual Deep Q-Network} 
While the aforementioned FS algorithm effectively reduces the communication load by semantically filtering the reconstruction message, there is still room for further improvement through exploiting the temporal features of the reconstruction message.
Unlike the traditional framework that transmits the message $\hvec{m}_t$ at the start of each time slot, we propose a PPDQN algorithm deployed at the physical robot side to dynamically adjust the transmission interval, so that the semantic messages $\hvec{m}_t^*$ produced by the FS algorithm are only sent during specific time slots to reduce the temporal redundancies. Specifically, the PPDQN algorithm should learn to discard messages with minimal impact on reconstruction error. For instance, messages are less informative and can be transmitted less frequently when the velocity of robotic arm is low, as their resulting movement (angle changes) is very limited. 
\subsubsection{Constrained Partially
Observable Markov Decision Process (C-POMDP) Problem}
We aim to dynamically optimise the number of transmission times of the reconstruction messages, with the objective to solve the error-constrained communication load minimisation problem formulated in Eq. (\ref{equ:problem}). Firstly, we adopt the Lagrangian primal-dual optimisation method and introduce Lagrange multipliers to incorporate the error constraints into the optimisation problem, the dual-problem is then formulated as
\begin{align} 
    \min_{\lambda^1, \lambda^2}\max_\pi  & \sum_{k=t}^\infty \gamma^{k-t} \mathbb{E}_\pi [-L_k] - \lambda^1 (C_{q_t} - e_{q_t}) - \lambda^2 (C_{\dot{q}_t} - e_{\dot{q}_t}),
\end{align}
where $\lambda^1$ and $\lambda^2$ are the Lagrange multipliers corresponding to the error constraints on the joint positions and joint velocities, $\gamma \in [\, 0,1)$ is the discount factor, and $\pi$ is the policy.

Note that the agent deployed at the physical robot does not have prior knowledge of whether the message being transmitted at the current moment will be successfully decoded by the receiver until the corresponding ACK packet is received. This uncertainty prevents the agent from fully observing state transitions, and makes our problem a C-POMDP, whose key components are given as follows:
\begin{itemize}
\item{\textbf{State:}} The state at the $t$-th time slot $S_t =\{ \mathcal{Q}_t, \dot{\mathcal{Q}}_{t}, \delta^c_{t-1} \}$ contains three parts: the joint angle set $\mathcal{Q}_t$, the joint velocity set $\dot{\mathcal{Q}}_{t}$, and the success or failure transmission of the previous message $\delta^c_{t-1}$. Before being fed into the neural network, both the joint angles and velocities must be scaled using min-max normalisation, where each value is divided by the robot joint angle range or velocity range, to mitigate the difference in input scales and enhance generalisation performance.
\item{\textbf{Action:}} The agent's action at the $t$-th time slot $A_t = \{0, 1\}$ represents the decision of whether to transmit each message, where 0 indicates discarding the message and 1 represents transmitting it.
\item{\textbf{Reward:}} The reward at the $t$-th time slot $r_t$ is denoted as the normalised communication load produced by transmitting the current message, which is expressed as 
\begin{align}
r_t = - \frac{L_t}{L_t^\text{th}},
\end{align}
where $L_t^\text{th}$ is the communication load produced by the traditional reconstruction framework in Sec. \ref{sec:3}. Note that $L_t=0$ if the agent chooses to discard the message.
\item{\textbf{Cost:}} The agent is expected to reduce the communication load as much as possible subject to the reconstruction error constraint, therefore the cost at the $t$-th slot $C_t$ is denoted as the current reconstruction error using
\begin{align}
    C_t = \lambda^1 e_{q_t} + \lambda^2 e_{\dot{q}_t}.
\end{align}
\end{itemize}

To tackle the intractable C-POMDP problem, we propose to integrate the DQN algorithm with the Lagrangian primal-dual optimisation method, and use PID controllers to obtain the optimal dual variables that correspond to the optimal policy. This method, known as PPDQN algorithm, is summarised in \textbf{Algorithm \ref{alg:DQN}.}
\begin{algorithm}[t]
    \caption{PID-based primal-dual Deep Q-Network} \label{alg:DQN}
    \textbf{Initialisation:} Training episode $M$, Replay memory $D$,  Q-network with parameters $\boldsymbol{\theta}$ and target network with parameters $\boldsymbol{\theta}^-$, Lagrange Multiplier $\lambda$
    \begin{algorithmic}
    \For{$j=1$ to $M$}
    \For{$i=1$ to $N_T$}
    \If{with probability $\epsilon$} 
    \State Select a random action $A_t \in \mathcal{A}$, 
    \Else
    \State Select $A_t=\arg\max_{A \in \mathcal{A}}Q(S_t,A;\boldsymbol{\theta}_t)$.
    \EndIf
    \State Execute action $A_t$, observe reward $r_t$ and cost $C_t$
    \State Achieve the next state $S_{t+1}$
    \State Store the state transition $(S_{t},A_{t},R_{t}, S_{t+1})$ in replay memory $D$
    \State Randomly sample mini-batch of transitions from replay memory $D$
    \State Perform a gradient descent step to minimise the loss using Eq. (\ref{equ:loss2})
    \State Update Q-network $\boldsymbol{\theta}$ using Eq. (\ref{equ:theta})
    \State Update target network $\boldsymbol{\theta}^-$ every $N_\theta$ steps
    \EndFor
    \If{not max episode}
    \State Update $\lambda^1$ and $\lambda^2$ using Eq. (\ref{equ:PID})
    \EndIf
    \EndFor
    \end{algorithmic}
\end{algorithm}
\subsubsection{Deep Q-Network} 
The DQN algorithm maintains two deep neural networks with the same architecture, a Q-network with parameter matrix $\boldsymbol{\theta}$ and a target network with parameter matrix $\boldsymbol{\theta}^-$, to avoid the learning instability caused by non-stationary target values. At each training step, the current state $S_t$ and its corresponding action $A_t$ are fed to the Q-network to calculate the Q-value (i.e., the cumulative reward of tasking action $A_t$ for the state $S_t$ under policy $\pi$) using the state-action function $Q(S_t, A_t;\boldsymbol{\theta}_t)$.
Note that the state action value in our network is taken with respect to both the reward and cost, which is expressed as
\begin{align}
Q(S_t, A_t) =R_{t}+\gamma \mathop {\text {max}}\limits _{A \in \mathcal{A}} Q(S_{t+1},A;\boldsymbol {\theta}_t) - C_t. 
\end{align}
Meanwhile, the target network takes the next state $S_{t+1}$ as input to estimate the maximum Q-value for $S_{t+1}$ over all possible actions. Intuitively, the goal is to train the agent to always select the action that maximises the long-term reward.
Therefore, the outputs of these two networks are used for minimising the loss function using Stochastic Gradient Descent. The loss function is given by 
\begin{align} L(\boldsymbol {\theta }_i) =  \mathbb {E}_{S_{t},A_{t},R_{t}, S_{t+1}} \big [ \big(R_{t}  &+  \gamma \mathop {\text {max}}\limits _{A \in \mathcal{A}} Q(S_{t+1},A;\boldsymbol {\theta}^-_t) \nonumber \\ & -Q(S_{t},A_{t};\boldsymbol {\theta}_t)\big )^{2} \big ],  \label{equ:loss1}
\end{align}
and its gradient can therefore be expressed as
\begin{align} \nabla L(\boldsymbol {\theta }_t)= &{\mathbb E}_{S_{t},A_{t},R_{t}, S_{t+1}}  \big [\big (R_{t+1}+  \gamma \mathop {\text {max}}\limits _{A \in \mathcal{A}}Q(S_{t+1}, A; {\boldsymbol {\theta }}_{t}^-)  \nonumber \\ & \qquad
-Q(S_{t}, A_{t}; \boldsymbol {\theta }_{t}) \big) \nabla _{\boldsymbol {\theta }} Q(S_{t}, A_{t}; \boldsymbol {\theta }_{t})\big]. \label{equ:loss2}
\end{align}
In this way, the parameter matrix of the Q-network is updated towards minimising the difference between the predicted Q-values and the target Q-values using the RMSProp optimiser 
\begin{align} {\boldsymbol \theta }_{t+1} = {\boldsymbol \theta }_{t} - \lambda _{\text {RMS}} \nabla L({\boldsymbol \theta }_{t}), \label{equ:theta}
\end{align}
in which $\lambda_{\text {RMS}}$ is the learning rate. Meanwhile, the parameter matrix of target network ${\boldsymbol \theta}^-$ is updated by copying the weights from the Q-network every $N_{\theta}$ steps to address the oscillations and divergence of the target values. 

Also note that the DQN algorithm stores the agent's experience tuple $(S_{t},A_{t},R_{t}, S_{t+1})$ at every training step in the replay memory $D$. During the training process, the algorithm randomly samples a mini-batch of experiences from the replay memory to break the temporal correlation between consecutive experiences. The expectations in Eq. (\ref{equ:loss1}) and Eq. (\ref{equ:loss2}) are both taken with respect to the mini-batch.


To achieve a balance between exploration and exploitation, we adopt an $\epsilon$-greedy approach for action selection. At each training step, the agent selects a random action with probability $\epsilon$ to explore the environment. Otherwise, it chooses the action that maximises the reward based on the current Q-value table. The probability $\epsilon$ is initially set to a high value and gradually decreases. Such a greedy approach ensures sufficient environment exploration to avoid local optima, while progressively increasing exploitation of the most rewarding actions to facilitate convergence.

\subsubsection{PID Lagrangian Method}
In this work, we use the Lagrangian method to address the constrained optimisation problem. 
The Lagrange multiplier $\lambda$ is introduced to serve as a penalty coefficient which balances the trade-off between minimising the communication load and satisfying the reconstruction error constraints. 
The classical Lagrangian Method increases the Lagrange multiplier $\lambda$ to amplify the penalty on the cost function when constraints are violated, and reduces it for satisfied constraints. Specifically, the update of multipliers $\lambda^1$ and $\lambda^2$ occurs every episode under a so-called integral control, which is expressed as 
\begin{align}
    &\lambda_j^1 = \max\left ( \, \lambda_{j-1}^1 + (\bar e_{q_j} - \bar  C_{{q}_j}), \, 0 \, \right ), \\
    &\lambda_j^2 = \max\left ( \, \lambda_{j-1}^2 + (\bar e_{\dot{q}_j} - \bar  C_{\dot{q}_j}), \, 0 \, \right ),
\end{align}
in which $j$ is the index of the training episode, $\bar e_{q_j}$ and $\bar e_{\dot{q}_j}$ are the average normalised joint angle error and velocity error in the $j$-th episode, $\bar  C_{{q}_j}$ and $\bar  C_{\dot{q}_j}$. are the corresponding average error constraints.
However, the classical Lagrangian method frequently leads to cost overshoot and oscillations, leading to instability in the training process and difficulty in satisfaction of the constraints. In our considered scenario, this problem would cause the algorithm to miss the optimal policy and converge before achieving the minimum communication load.

To tackle this, we introduce two PID controllers \cite{PID} to update the Lagrange multipliers $\lambda^1$ and $\lambda^2$, respectively. Let us take the update of $\lambda^1$ as an example, which is used to penalise the joint angle error. We incorporate the proportional control and the derivative control alongside the original integral control, in which the update rule is expressed as
\begin{align} 
    \lambda_j^1 =  \max\left ( \, K_P \Delta_j^1 + K_I I_j^1 + K_D D_j^1, \, 0 \, \right ), \label{equ:PID}
\end{align}
where $K_P$, $K_I$ and $K_D$ are the proportional, integral, and derivative gains, respectively. $\Delta_j^1$, $I_j^1$, and $D_j^1$ are the proportional, integral, and derivative errors, which are obtained by
\begin{align}
    &\Delta_j^1 = \bar e_{q_j} - \bar C_{q_j}, \label{equ:PID1} \\
    &I_j^1 = \max\left (\, I_{j-1}^1 + \Delta_j^1, \, 0 \right ), \label{equ:PID2} \\
    &D_j^1 = \max\left (\, \bar e_{q_j} - \bar e_{q_{j-1}}, \, 0 \right ). \label{equ:PID3}
\end{align}
The proportional control term $K_P \Delta_j^1$ in Eq. (\ref{equ:PID}) is updated using Eq. (\ref{equ:PID1}), which characterises how much the constraint is violated. It is directly proportional to the constraint violation, and thus provides an immediate correction based on the current reconstruction error to rapidly mitigate any large error violations. The integral control term $K_I I_j^1$, as described in Eq. (\ref{equ:PID2}), accumulates the previous reconstruction errors over time, and continuously penalises the system until the reconstruction error constraint is satisfied, so that even a small, yet continuous error will be accumulated and eventually eliminated. The derivative control term $K_D D_j^1$ is calculated by taking the difference between the average error at the current episode and the last episode, which reflects the reconstruction error change rate. When overshoot and oscillations occur, its value increases immediately in response to the rapid change in the reconstruction error, thereby dampening these effects.
Compared with the classical Lagrangian Method, the introduction of proportional control and derivative control allows for the fine-tuning of the Lagrange multiplier $\lambda$, resulting in not only a faster response to error violations and overshoot but also an effective mitigation of error oscillations after convergence.

\section{Simulation Results}  \label{sec:sim}
In this section, we validate the effectiveness of our proposed GSC reconstruction framework and compare it with the traditional framework via both simulations and experiments.

\subsection{System Setup}
The robot arm considered in the simulations and experiments is the 7-DoF Franka Research 3 (FR3) equipped with a two-finger parallel gripper. Its low-level control frequency is preset by the manufacturer at 1 kHz, hence the message transmission interval is set to be 1 ms. 
The settings for the channel parameters and hyperparameters are given in Table \ref{tab:1}. 
Please note that the transmit power $P$ is chosen to represent a typical low-power wireless transmitter, such as WiFi and ZigBee. The SNR threshold $\beta$ is selected to simulate WiFi sensing capability constraint. These parameters along with the noise power follow the standard network settings specified in \cite{channel}. The distance is chosen to simulate a large-scale industrial factory setup, where robotic arms may be located at significant distances from the edge server. Also note that the Nakagami-$m$ fading model is simplified to Rayleigh fading in the simulations.
\subsubsection{Simulation Design}
We reconstruct the real-time pick-and-place, pick-and-toss and push-and-pull tasks in the PyBullet simulator \cite{pybullet}. For the pick-and-place task, the robot arm is designed to move a wooden cube from its initial position to a specified target location. In the pick-and-toss task, the robot arm must throw the wooden cube accurately into a target area. For the push-and-pull task, the robot arm is required to open a drawer and then close it. 
The digital model of the FR3 is provided by its manufacturer, and aligns with the size, mechanical structure, and kinematics of the real FR3 robot arm. The simulator enables real-time monitoring of the robot's states and interaction, these data are fed to the DRL agent for online training.
All simulations were conducted on a single workstation, with each DT employing different reconstruction frameworks and communicating with the same simulated physical robot (serving as the physical twin) via the same PyBullet simulation. 
We consider two communication cases:
(i) a virtual ideal wired link, which is represented by perfect and lossless message transmissions in simulations,
(ii) a virtual wireless channel, which is digitally modelled at the transmitter side to mimic real-world channel conditions and determines the uplink transmission from the physical robot to the DTs. In wireless cases, the simulated physical robot  transmits reconstruction messages to multiple DTs under the same channel conditions simultaneously to ensure a fair comparison across baselines since all configurations on robot and communication setup are the same.

\subsubsection{Experimental Design}
\begin{figure}
\centering
\includegraphics[width=1\linewidth]{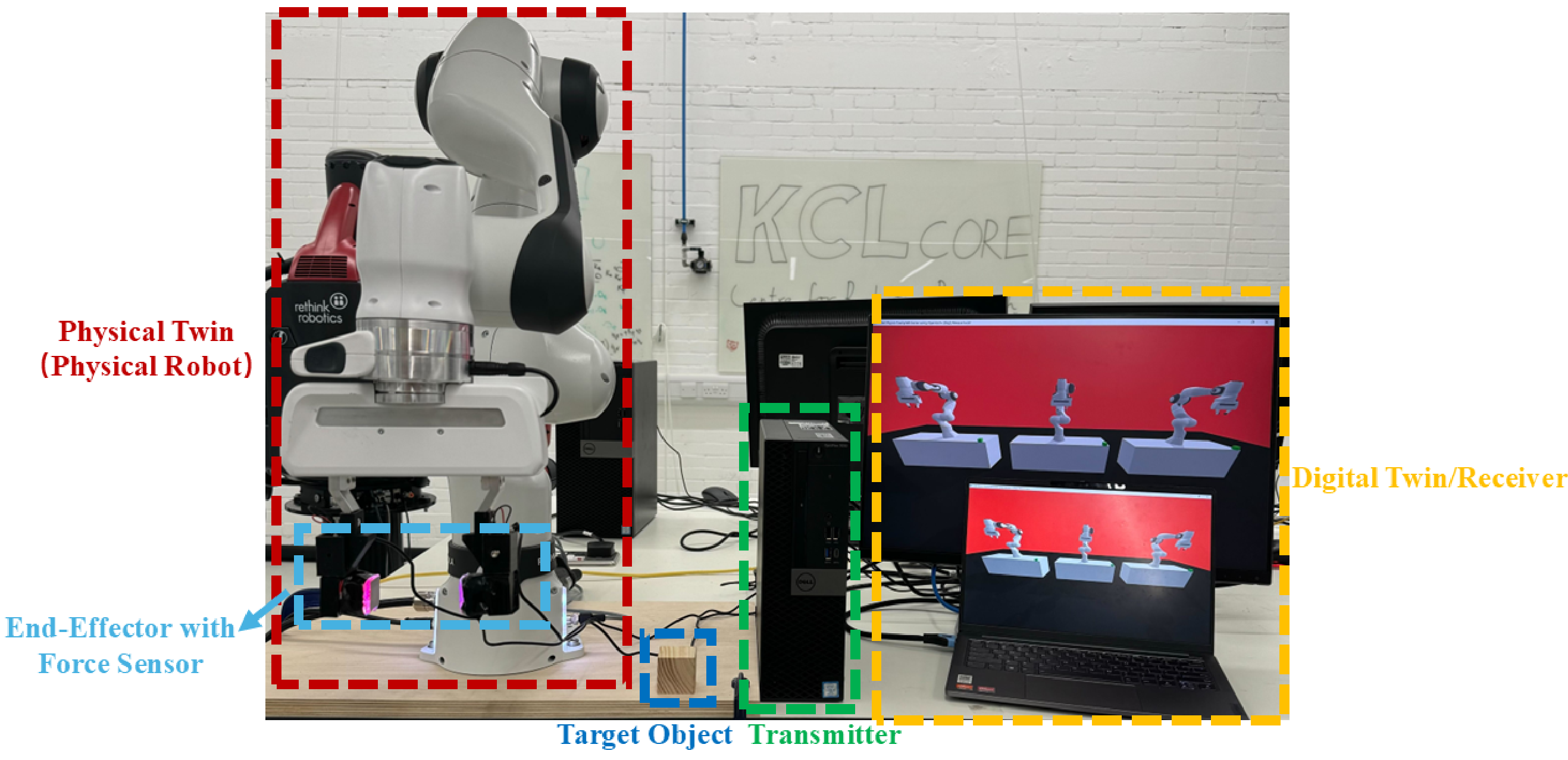}
\caption{Real-world testbed setup.}
\label{fig:testbed}
\end{figure}

We also conduct real-world experiments for the pick-and-place task, where the testbed setup is shown in Fig. \ref{fig:testbed}. 
The physical robot arm is commanded by an external workstation PC (i.e., the transmitter in Fig. \ref{fig:testbed})  via the C++ interface \emph{libfranka} at 1 kHz, and the digital robot arm is deployed in another workstation PC running a PyBullet simulation. Both the physical robot arm and the workstations are physically connected to the same local area network via Gigabit Ethernet, which can be considered as a perfect transmission medium with near-zero latency and packet loss.
In this setup, the physical robot arm executes the commands and sends back reconstruction messages to its connected workstation PC. These reconstruction messages then undergo both feature and temporal selections at the transmitter side, and are forwarded to the other workstation that hosts multiple DTs running different reconstruction frameworks through Ethernet.  Similar to simulations, both wired and wireless communication links are compared, where the Ethernet can be regarded as an idea wired link, while
a simulated wireless channel is introduced to the forward link to mimic the practical wireless channel, which employs the same parameter settings as the simulation to ensure testing condition consistency when comparing experimental results with simulation outcomes.

To capture the contact force, a Gelsight sensor \cite{gelsight} is attached to the surface of the gripper pads, which is a gel-based tactile sensor that can perceive the shape change of the contact surface and identify 3-axis contact forces. Comparisons of both communication load and reconstruction error between the baselines and our proposed framework are illustrated in Fig. \ref{fig:e}.

\subsubsection{Baselines}
We compare the following frameworks in both simulations and experiments:
\begin{itemize}
\item \textbf{Traditional Wired Framework} (labelled as ‘‘Wired”): the physical robot transmits the complete message $\hvec{m}_t$ in Eq. (6) at every time step through wired link, which is achieved by a simulated high-bandwidth and error-free  wired channel in simulation and physical Ethernet connection in experiments. That is, both uplink and downlink transmissions are ideal with the simulated wireless channel deactivated.
\item \textbf{Traditional Wireless Framework} (labelled as ‘‘Wireless”): the physical robot transmits the complete message $\hvec{m}_t$  at every time step via the virtual wireless link, which means the uplink channel is non-ideal with Nakagami-$m$ fading but downlink channel is still ideal.
\item \textbf{GSC Framework with Feature Selection but without Temporal Selection} (labelled as ‘‘FS”): the physical robot transmits the semantic information $\hvec{m}_{t}^*$  at every time step via wireless link.
\item \textbf{Our Proposed GSC Framework} (labelled as ‘‘FS\&PPDQN”): the physical robot transmits the semantic information $\hvec{m}_{t}^*$  at selected time steps via wireless link.
\end{itemize}

Additionally, we present the performance of our proposed GSC framework under both strict and relaxed reconstruction error constraints to demonstrate its adaptability in balancing the communication load and the reconstruction error:
\begin{itemize} 
\item \textbf{Strict Constraint} (labelled as ‘‘SC”): A strict error constraint is applied to the GSC framework, which allows the average normalised angle error to increase by up to 0.002\% and the average normalised velocity error to increase by up to 0.005\% as compared to the Traditional Wireless Framework. 
\item \textbf{Relaxed Constraint} (labelled as ‘‘RC”): The GSC framework is allowed to tolerate a higher reconstruction error in exchange for a reduced communication load in the relaxed constraint case, where the average normalised angle error is allowed to increase by up to 0.02\% while the average normalised velocity error can increase by up to 0.05\% compared to the Traditional Wireless Framework.
\end{itemize}

\begin{table}[]
\centering
\caption{Channel parameters and hyperparameters}
\renewcommand\arraystretch{1.2}
\label{tab:1}
\begin{tabular}{|cccc|}
\hline
\multicolumn{4}{|c|}{\textbf{Channel Parameters}} \\ \hline
\multicolumn{1}{|c|}{Shape parameter $m$} & \multicolumn{1}{c|}{1} & \multicolumn{1}{c|}{Scale parameter $\Omega$} & 1 \\ \hline
\multicolumn{1}{|c|}{Path loss coefficient $\alpha$} & \multicolumn{1}{c|}{4.31} & \multicolumn{1}{c|}{Distance $d$} & 110m \\ \hline
\multicolumn{1}{|c|}{Transmit power $P$} & \multicolumn{1}{c|}{{18dBm}} & \multicolumn{1}{c|}{Noise power $\sigma^2$} & -90dBm \\ \hline
\multicolumn{1}{|c|}{SNR threshold $\beta$} & \multicolumn{1}{c|}{{5.5dB}} & \multicolumn{1}{c|}{Time interval} & 1ms \\ \hline
\multicolumn{4}{|c|}{\textbf{Hyperpameters}} \\ \hline
\multicolumn{1}{|c|}{Replay memory size $D$} & \multicolumn{1}{c|}{40000} & \multicolumn{1}{c|}{Mini-batch size} & 32 \\ \hline
\multicolumn{1}{|c|}{Discount rate $\gamma$} & \multicolumn{1}{c|}{0.9} & \multicolumn{1}{c|}{Learning rate} & $10^{-5}$ \\ \hline
\multicolumn{1}{|c|}{Initial exploration $\epsilon$} & \multicolumn{1}{c|}{0.99} & \multicolumn{1}{c|}{Final exploration $\epsilon$} & 0.001 \\ \hline
\multicolumn{1}{|c|}{Optimizer} & \multicolumn{1}{c|}{RMSProp} & \multicolumn{1}{c|}{Activation function} & ReLU \\ \hline
\end{tabular}
\end{table}

\begin{figure*}[htb]
    \centering
    \begin{subfigure}[b]{0.33\linewidth}
        \includegraphics[width=\linewidth]{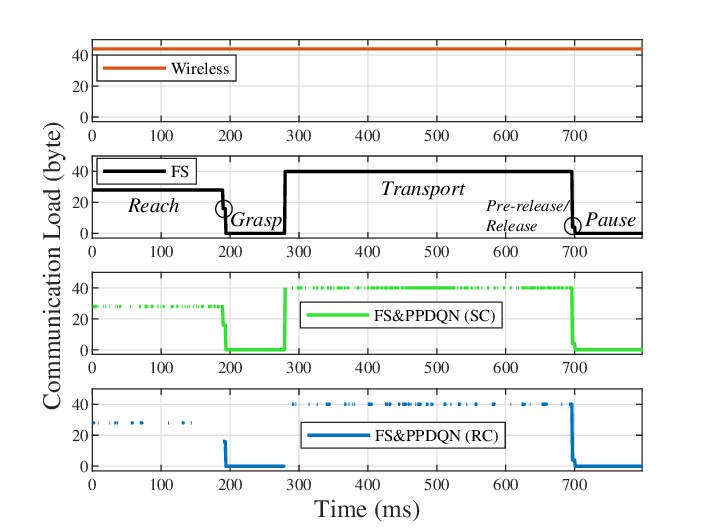}
        \caption{Comparison of instant communication load}
        \label{fig:s_place_load}
    \end{subfigure}
    \hspace{-0.03\linewidth}
    \begin{subfigure}[b]{0.32\linewidth}
        \includegraphics[width=\linewidth]{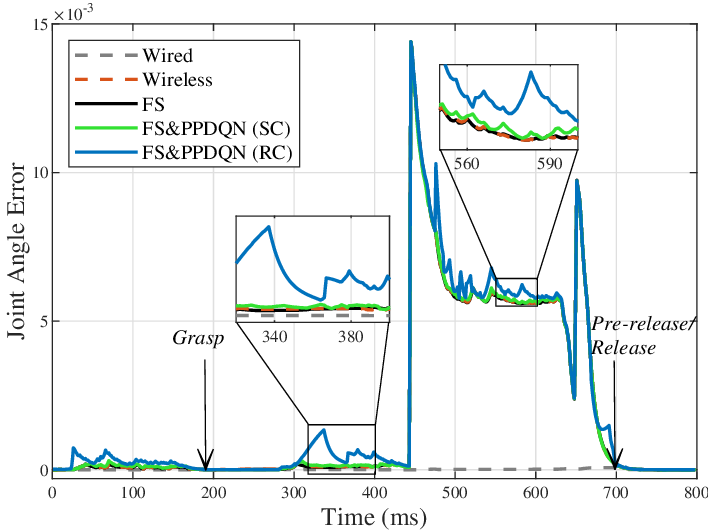}
        \caption{Comparison of normalised joint angle error}
        \label{fig:s_place_angle}
    \end{subfigure}
    \begin{subfigure}[b]{0.32\linewidth}
        \includegraphics[width=\linewidth]{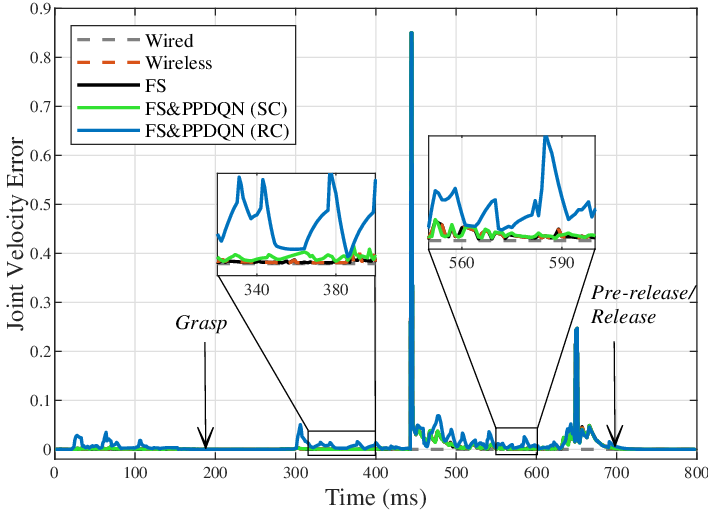}
        \caption{Comparison of normalised joint velocity error}
        \label{fig:s_place_vel}
    \end{subfigure}
    \caption{\small Performance comparison for pick-and-place task between baselines and the proposed GSC framework in simulations.}
    \label{fig:s_place}
    \vspace{-0.3cm}
\end{figure*}\vspace{-0.2cm}
\subsection{Performance Comparison in Simulations}
\begin{figure*}[htb]
    \centering
    \begin{subfigure}[b]{0.33\linewidth}
        \includegraphics[width=\linewidth]{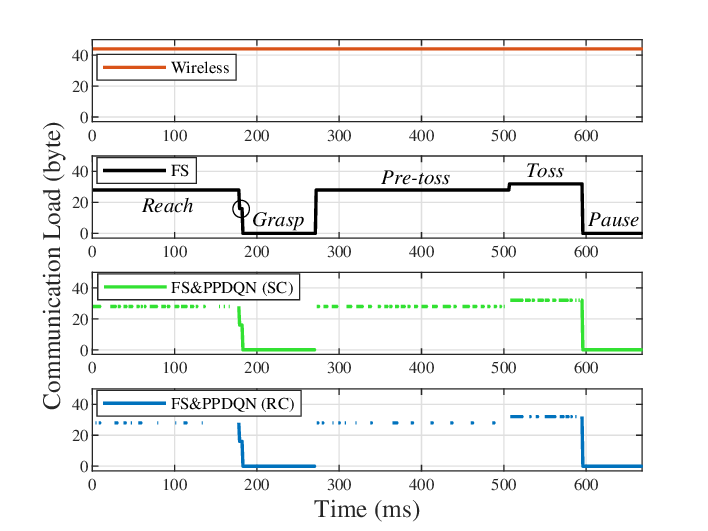}
        \caption{Comparison of instant communication load}
        \label{fig:s_toss_load}
    \end{subfigure}
    \hspace{-0.03\linewidth}
    \begin{subfigure}[b]{0.32\linewidth}
        \includegraphics[width=\linewidth]{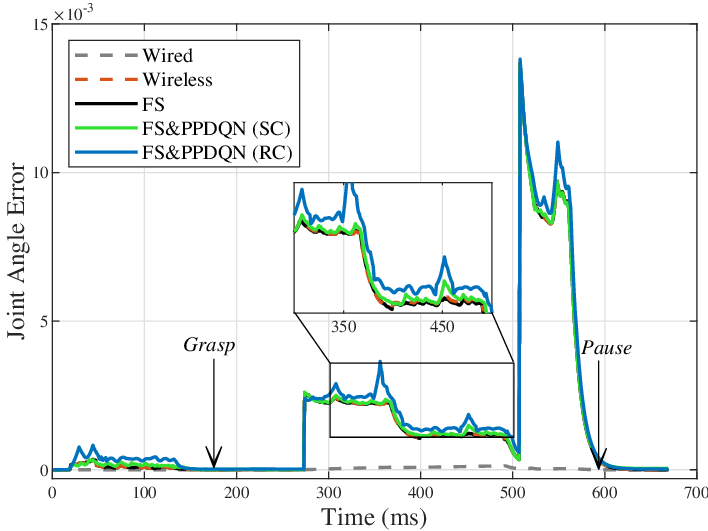}
        \caption{Comparison of normalised joint angle error}
        \label{fig:s_toss_angle}
    \end{subfigure}
    \begin{subfigure}[b]{0.32\linewidth}
        \includegraphics[width=\linewidth]{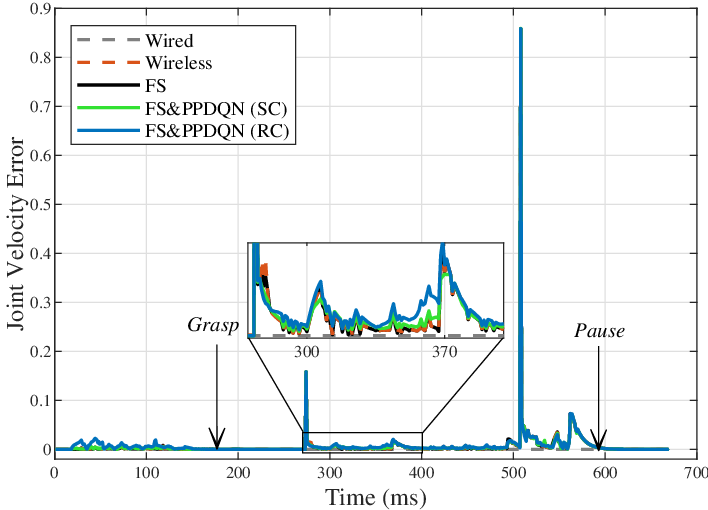}
        \caption{Comparison of normalised joint velocity error}
        \label{fig:s_toss_vel}
    \end{subfigure}
    \caption{\small Performance comparison for pick-and-toss task between baselines and the proposed GSC framework in simulations.}
    \label{fig:s_toss}
    \vspace{-0.3cm}
\end{figure*}

We compare the communication load and the reconstruction error of baselines and our GSC framework for the pick-and-place task and pick-and-toss task in Fig. \ref{fig:s_place} and Fig. \ref{fig:s_toss}, respectively. Note that the results of the push-and-pull task are omitted here as they are similar to those of the pick-and-place task, but they can be found in the provided demo. 

Fig. \ref{fig:s_place_load} plots the instantaneous communication load of the Traditional Wireless Framework and our GSC framework versus time for the pick-and-place task in the simulator, where missing dots at specific times represent time slots with no transmission occurring. Compared with the Traditional Wireless Framework (top subfigure) which maintains a constant communication load of 44 bytes, the FS algorithm (second subfigure) can dynamically adjust the transmitted features and significantly decrease the communication load according to the current phase at every time slot. Moreover, the PPDQN algorithm (bottom two subfigures) further reduces the communication load due to the reduced number of transmissions. This shows our GSC framework only requires transmitting a much lower number of bytes. Importantly, compared to the relaxed constraint case (fourth subfigure), the number of transmission times in the strict constraint case (third subfigure) is much denser with more points. This indicates that in the strict constraint case, transmissions occur more frequently to meet the stringent requirements in the reconstruction error, resulting in a correspondingly higher communication load. 
Notably, the communication load drops to 0 after $t=190$ ms and $696$ ms (bottom three subfigures) due to our semantics-aware transmission strategy of gripper state, where the physical robot discards subsequent gripper state messages that contain the same information once the previous one is successfully delivered.

Fig. \ref{fig:s_place_angle} and Fig. \ref{fig:s_place_vel} plot the normalised joint angle error and normalised joint velocity error over time under baselines and our proposed GSC framework when performing the pick-and-place task. We first observe that the angle error and the velocity error of the GSC framework are both relatively close to those of the Traditional Wireless Framework.
This indicates that our framework did not introduce any additional errors while effectively reducing the communication load. 
We also observe that the errors produced by Traditional Wired Framework with ideal reconstruction message transmission are negligible compared to the rest wireless-based frameworks, which shows that the reconstruction errors are mainly caused by the unstable wireless channel.
Meanwhile, it can be seen that the reconstruction error increases when its constraints are relaxed as messages that contribute less to improving the reconstruction accuracy are discarded. 
Importantly, combined with Fig. \ref{fig:s_place_load}, one can see that the transmission interval (i.e., the density of points) is dynamically adjusted according to the reconstruction error, with more messages being transmitted when the reconstruction error is larger.
Also, it can be seen that the error performance of both the Traditional Wireless Framework and our GSC framework significantly deteriorate when the robot is in contact with the object (i.e., the period between the \textit{grasp} phase and the \textit{release} phase). This is because the interaction magnifies the errors caused by the wireless channel. On the one hand, the packet loss, especially the loss of contact force, during this period will affect the robot’s dynamics according to Eq. \ref{equ:tau}, which may in turn cause the digital robot to apply insufficient torque or overcompensate for external forces, ultimately disrupting its own behaviour. On the other hand, the transmission failure of joint angles before the \textit{grasp} phase could result in position deviation of the contact point between the gripper and the object, where this inconsistent contact status might cause error escalation and accumulation. This again highlights the novelty of our GSC framework because prior research \cite{9530501,9953092,10370739} always considered non-contact scenarios, and neglected the compounded effects of the unstable wireless channel and interactions that can potentially degrade the reconstruction quality.

\begin{figure*}[htb]
\centering
    \begin{subfigure}[b]{0.5\linewidth}
        \includegraphics[width=\linewidth]{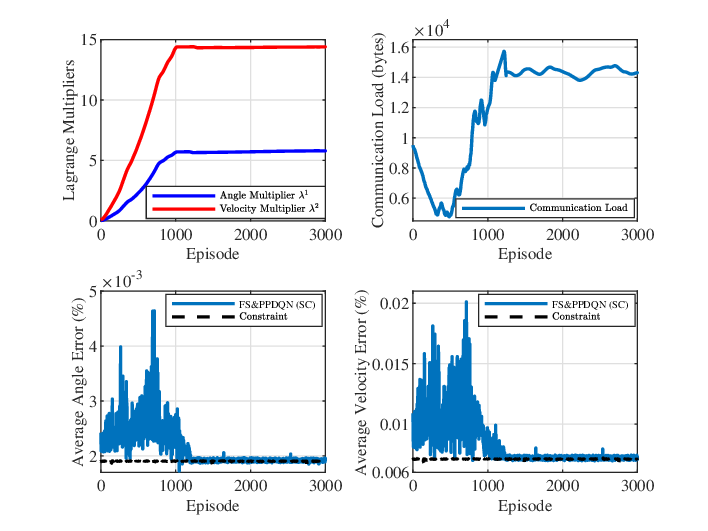}
        \setlength{\abovecaptionskip}{-15pt}
        \caption{Strict constraint case}
        \label{fig:strict_case}
    \end{subfigure}
    \hspace{-0.02\linewidth}
    \begin{subfigure}[b]{0.5\linewidth}
        \includegraphics[width=\linewidth]{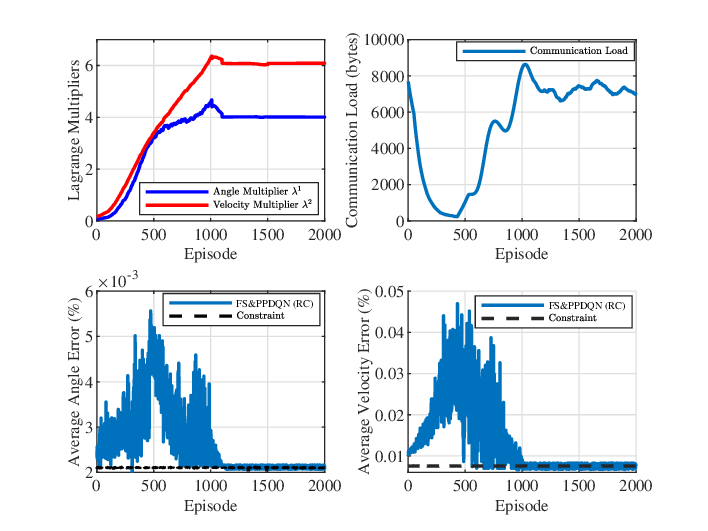}
        \setlength{\abovecaptionskip}{-15pt}
        \caption{Relaxed constraint case}
        \label{fig:relax_case}
    \end{subfigure}
    \caption{\small Training performance evaluation for pick-and-place task with PID controller.}
    \label{fig:PID}
    \vspace{-0.3cm}
\end{figure*}
Fig. \ref{fig:s_toss_load} plots the communication load of the Traditional Wireless Framework and our proposed GSC framework for the pick-and-toss task in simulation. Similarly, it is observed that the FS algorithm can reduce the communication load in the feature domain via only transmitting the GSC information, and the PPDQN algorithm adaptively adjusts the transmission frequency based on the reconstruction accuracy requirement.

Fig. \ref{fig:s_toss_angle} and Fig. \ref{fig:s_toss_vel} present the real-time reconstruction error under baselines and our proposed GSC framework for the pick-and-toss task.
Firstly, we found that our GSC framework achieves comparable reconstruction accuracy to the Traditional Wireless Framework.
Secondly, we see that the PPDQN algorithm is capable of satisfying different error constraints based on its learned policy. 
Thirdly, we also see that the reconstruction error significantly increases in contact phases, which again revealing the compounded effect of the packet loss and interaction.

\subsection{Training Performance with PID-based Lagrangian Method}
Our proposed PPDQN algorithm aims to dynamically adjust the value of the Lagrange Multiplier based on the DT reconstruction error and the preset error constraint during the training process. Herein, we take the training for the pick-and-place task in both strict constraint and relaxed constraint cases as an example, where the variations of the Lagrange Multiplier, cumulative communication load and reconstruction error performance are illustrated in Fig. \ref{fig:PID}. 

In the early stages of training, the agent explores the environment extensively, during which the communication load gradually decreases in both cases. This reduction leads to an increase in reconstruction error since less messages are transmitted. Consequently, the DT keeps violating the error constraint, and thus drives the PID controller to increase the values of the multipliers. This imposes a greater penalty on the reconstruction error and encourages the agent to transmit more messages to meet the constraints. Notably, the increasing rate of the multiplier is positively correlated with the disparity between the current reconstruction error and the error constraint. Once the error constraints are satisfied, the multiplier stops increasing and the agent starts to minimise the communication load. Finally, the system reaches the steady-state when the optimal policy is achieved, with only minor adjustment being made in response to reconstruction error oscillation.

It can also be observed that the Lagrange multipliers grow more rapidly under strict error constraints, as shown in Fig. \ref{fig:strict_case}. This is because the proportional term reacts more intensely to large deviations caused by the stringent constraints, and the integral term accumulates the sustained errors more quickly. This drives the cost to escalate rapidly, which prompts the agent to increase transmission frequency. Consequently, the cumulative communication load increases sharply before stabilizing at approximately 14 kB.
In contrast, as shown in Fig. 7(b), the Lagrange multipliers grow more gradually under relaxed constraints. This leads to a slower escalation of both the cost and the communication load. Consequently, the cumulative communication load coverages at a lower value of approximately 7.2 kB.

\begin{figure*}[htb]
    \centering
    \begin{subfigure}[b]{0.33\linewidth}
        \includegraphics[width=\linewidth]{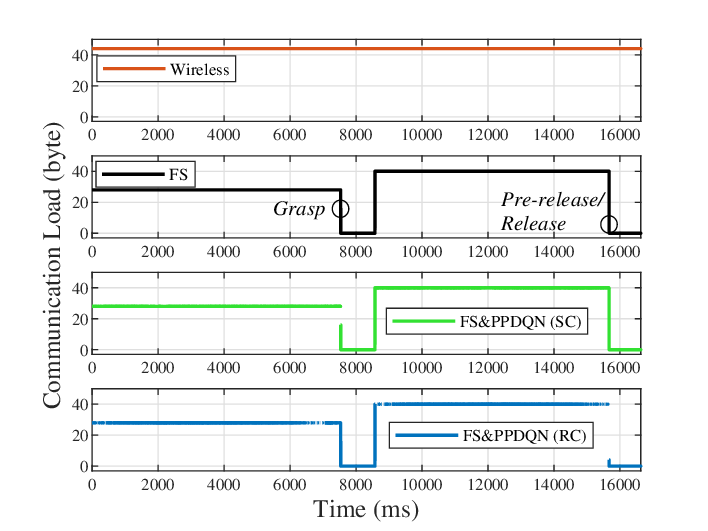}
        \caption{Comparison of instant communication load}
        \label{fig:e_load}
    \end{subfigure}
    \hspace{-0.03\linewidth}
    \begin{subfigure}[b]{0.315\linewidth}
        \includegraphics[width=\linewidth]{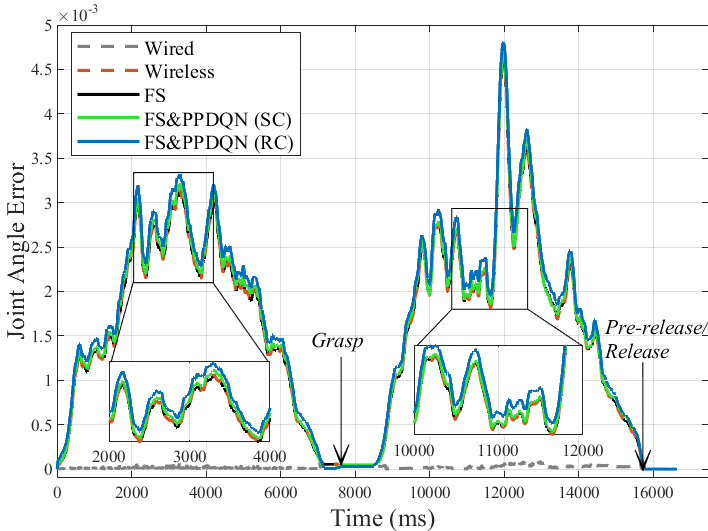}
        \caption{Comparison of normalised joint angle error}
        \label{fig:e_angle}
    \end{subfigure}
    \begin{subfigure}[b]{0.315\linewidth}
        \includegraphics[width=\linewidth]{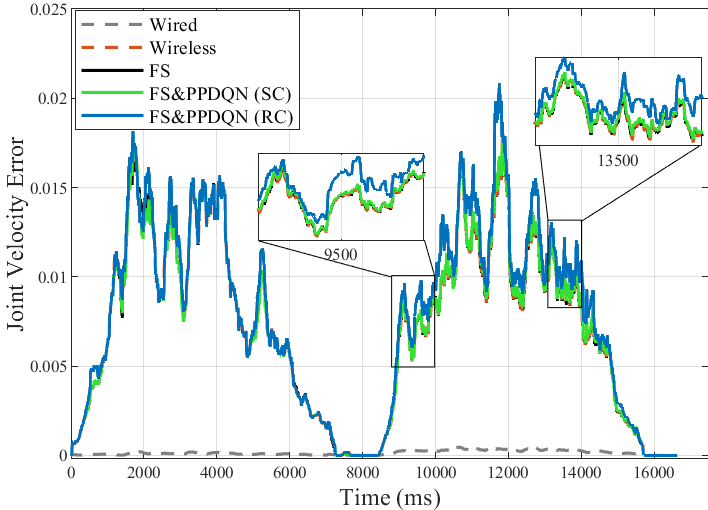}
        \caption{Comparison of normalised joint velocity error}
        \label{fig:e_vel}
    \end{subfigure}
    \caption{\small Performance comparison for pick-and-place task between baselines and the proposed GSC framework in experiments.}
    \label{fig:e}
\end{figure*}

\subsection{Runtime Evaluation}
Given that the DT reconstruction task operates in real-time within robotic systems, we also evaluate whether the proposed PPDQN algorithm can meet the timeliness constraints. We focus solely on the runtime evaluation during the deployment phase, as the expensive training process can be performed offline in advance. Therefore, we only consider the time for forward propagation in the deployment phase, and omit both forward and backward propagation, as well as gradient updates in the training phase.

The computational complexity of our proposed PPDQN algorithm during the deployment phase can be expressed as
$
O\left( n_s \cdot n_1 + \sum_{i=1}^{N_L-1} n_i \cdot n_{i+1} + n_{N_L} \cdot n_a \right),
$
where $n_s$ is the dimension of the input state, $n_a$ is the dimension of the action space, $n_i$ represents the number of units in the $i$-th layer, and $N_L$ is the  number of neural network layers.
We adopt a network architecture with two hidden layers, each consisting of $128$ units. The input state dimension is $n_s=15$, which consists of a 7-dimensional joint angle set, a 7-dimensional joint velocity set, and a 1-dimensional binary variable. The action space dimension is $n_a=1$, corresponding to the binary action. Therefore, the computational complexity of the proposed PPDQN algorithm is 18384 parameters.  In both simulation and experiment, the algorithm runs on a workstation equipped with a single NVIDIA RTX 3090 GPU, and the average runtime required to make an online decision is approximately 61 ns, which indicates that the proposed PPDQN algorithm is highly efficient in real-time system.

\begin{table*}[h]
\centering
\caption{\centering Overall performance comparison on different frameworks}
\label{tab:performance}
\renewcommand\arraystretch{1.2}
\begin{tabular}{|ccccccc|}
\hline
\multicolumn{7}{|c|}{\textbf{Pick-and-place (Simulations)}} \\ \hline
\multicolumn{1}{|c|}{Method} & \multicolumn{1}{c|}{\begin{tabular}[c]{@{}c@{}}Cumulative \\ Load (byte)\end{tabular}} & \multicolumn{1}{c|}{{\begin{tabular}[c]{@{}c@{}}Load \\ Reduction (\%)\end{tabular}}} & \multicolumn{1}{c|}{\begin{tabular}[c]{@{}c@{}}Average Angle \\ Error\end{tabular}} & \multicolumn{1}{c|}{{\begin{tabular}[c]{@{}c@{}}Angle Error \\ Increase (\%)\end{tabular}}} & \multicolumn{1}{c|}{\begin{tabular}[c]{@{}c@{}}Average Velocity \\ Error\end{tabular}} & {\begin{tabular}[c]{@{}c@{}}Velocity Error \\ Increase (\%)\end{tabular}} \\\hline
\multicolumn{1}{|c|}{Wired} & \multicolumn{1}{c|}{{35112}} & \multicolumn{1}{c|}{{N/A}} & \multicolumn{1}{c|}{$9.341 \times 10^{-6}$} & \multicolumn{1}{c|}{{N/A}} & \multicolumn{1}{c|}{$1.775 \times 10^{-6}$} & {N/A} \\
\multicolumn{1}{|c|}{Wireless} & \multicolumn{1}{c|}{35112} & \multicolumn{1}{c|}{{0}} & \multicolumn{1}{c|}{$1.909 \times 10^{-3}$} & \multicolumn{1}{c|}{{0}} & \multicolumn{1}{c|}{$7.020 \times 10^{-3}$} & {0} \\
\multicolumn{1}{|c|}{FS} & \multicolumn{1}{c|}{22052} & \multicolumn{1}{c|}{{37.2}} & \multicolumn{1}{c|}{$1.909 \times 10^{-3}$} & \multicolumn{1}{c|}{{0}} & \multicolumn{1}{c|}{$7.021 \times 10^{-3}$} & {0.0001} \\
\multicolumn{1}{|c|}{FS\&PPDQN(SC)} & \multicolumn{1}{c|}{14200} & \multicolumn{1}{c|}{{59.5}} & \multicolumn{1}{c|}{$1.926 \times 10^{-3}$} & \multicolumn{1}{c|}{{0.0017}} & \multicolumn{1}{c|}{$7.052 \times 10^{-3}$} & {0.0032} \\
\multicolumn{1}{|c|}{FS\&PPDQN(RC)} & \multicolumn{1}{c|}{7176} & \multicolumn{1}{c|}{{79.6}} & \multicolumn{1}{c|}{$2.108 \times 10^{-3}$} & \multicolumn{1}{c|}{{0.0199}} & \multicolumn{1}{c|}{$7.506 \times 10^{-3}$} & {0.0486} \\\hline
\multicolumn{7}{|c|}{\textbf{Pick-and-place (Experiments)}} \\ \hline
\multicolumn{1}{|c|}{Wired} & \multicolumn{1}{c|}{{731720}} & \multicolumn{1}{c|}{{N/A}} & \multicolumn{1}{c|}{$2.033 \times 10^{-5}$} & \multicolumn{1}{c|}{{N/A}} & $1.537 \times 10^{-6}$ & \multicolumn{1}{|c|}{{N/A}} \\
\multicolumn{1}{|c|}{Wireless} & \multicolumn{1}{c|}{731720} & \multicolumn{1}{c|}{{0}} & \multicolumn{1}{c|}{$1.551 \times 10^{-3}$} & \multicolumn{1}{c|}{{0}} & $7.212 \times 10^{-3}$ & \multicolumn{1}{|c|}{{0}} \\
\multicolumn{1}{|c|}{FS} & \multicolumn{1}{c|}{494864} & \multicolumn{1}{c|}{{32.3}} & \multicolumn{1}{c|}{$1.554 \times 10^{-3}$} & \multicolumn{1}{c|}{{0.0003}} & $7.218 \times 10^{-3}$ & \multicolumn{1}{|c|}{{0.0006}} \\
\multicolumn{1}{|c|}{FS\&PPDQN(SC)} & \multicolumn{1}{c|}{344584} & \multicolumn{1}{c|}{{52.9}} & \multicolumn{1}{c|}{$1.586 \times 10^{-3}$} & \multicolumn{1}{c|}{{0.0035}} & $7.259 \times 10^{-3}$ & \multicolumn{1}{|c|}{{0.0047}} \\
\multicolumn{1}{|c|}{FS\&PPDQN(RC)} & \multicolumn{1}{c|}{192904} & \multicolumn{1}{c|}{{73.6}} & \multicolumn{1}{c|}{$1.663 \times 10^{-3}$} & \multicolumn{1}{c|}{{0.0112}} & $7.749 \times 10^{-3}$ & \multicolumn{1}{|c|}{{0.0537}} \\\hline
\multicolumn{7}{|c|}{\textbf{Pick-and-toss (Simulations)}} \\ \hline
\multicolumn{1}{|c|}{Wired} & \multicolumn{1}{c|}{{29392}} & \multicolumn{1}{c|}{{N/A}} & \multicolumn{1}{c|}{$3.013 \times 10^{-5}$} & \multicolumn{1}{c|}{{N/A}} & $7.479 \times 10^{-7}$ & \multicolumn{1}{|c|}{{N/A}} \\
\multicolumn{1}{|c|}{Wireless} & \multicolumn{1}{c|}{29392} & \multicolumn{1}{c|}{{0}} & \multicolumn{1}{c|}{$1.480 \times 10^{-3}$} & \multicolumn{1}{c|}{{0}} & $5.978 \times 10^{-3}$ & \multicolumn{1}{|c|}{{0}} \\
\multicolumn{1}{|c|}{FS} & \multicolumn{1}{c|}{14476} & \multicolumn{1}{c|}{{50.8}} & \multicolumn{1}{c|}{$1.481 \times 10^{-3}$} & \multicolumn{1}{c|}{{0.0001}} & $5.980 \times 10^{-3}$ & \multicolumn{1}{|c|}{{0.0002}} \\
\multicolumn{1}{|c|}{FS\&PPDQN(SC)} & \multicolumn{1}{c|}{8712} & \multicolumn{1}{c|}{{70.4}} & \multicolumn{1}{c|}{$1.505 \times 10^{-3}$} & \multicolumn{1}{c|}{{0.0025}} & $6.129 \times 10^{-3}$ & \multicolumn{1}{|c|}{{0.0051}} \\
\multicolumn{1}{|c|}{FS\&PPDQN(RC)} & \multicolumn{1}{c|}{4060} & \multicolumn{1}{c|}{{86.2}} & \multicolumn{1}{c|}{$1.687 \times 10^{-3}$} & \multicolumn{1}{c|}{{0.0207}} & $6.501 \times 10^{-3}$ & \multicolumn{1}{|c|}{{0.0523}} \\\hline

\multicolumn{7}{|c|}{\textbf{Push-and-pull (Simulations)}} \\ \hline
\multicolumn{1}{|c|}{Wired} & \multicolumn{1}{c|}{{40700}} & \multicolumn{1}{c|}{{N/A}} & \multicolumn{1}{c|}{$6.405 \times 10^{-5}$} & \multicolumn{1}{c|}{{N/A}} & $2.157 \times 10^{-6}$ & \multicolumn{1}{|c|}{{N/A}} \\
\multicolumn{1}{|c|}{Wireless} & \multicolumn{1}{c|}{40700} & \multicolumn{1}{c|}{{0}} & \multicolumn{1}{c|}{$1.279 \times 10^{-3}$} & \multicolumn{1}{c|}{{0}} & $5.218 \times 10^{-3}$ & \multicolumn{1}{|c|}{{0}} \\
\multicolumn{1}{|c|}{FS} & \multicolumn{1}{c|}{25688} & \multicolumn{1}{c|}{{36.8}} & \multicolumn{1}{c|}{$1.276 \times 10^{-3}$} & \multicolumn{1}{c|}{{-0.0003}} & $5.215 \times 10^{-3}$ & \multicolumn{1}{|c|}{{-0.0003}} \\
\multicolumn{1}{|c|}{FS\&PPDQN(SC)} & \multicolumn{1}{c|}{15760} & \multicolumn{1}{c|}{{61.3}} & \multicolumn{1}{c|}{$1.295 \times 10^{-3}$} & \multicolumn{1}{c|}{{0.0016}} & $5.270 \times 10^{-3}$ & \multicolumn{1}{|c|}{{0.0052}} \\
\multicolumn{1}{|c|}{FS\&PPDQN(RC)} & \multicolumn{1}{c|}{7524} & \multicolumn{1}{c|}{{81.5}} & \multicolumn{1}{c|}{$1.499 \times 10^{-3}$} & \multicolumn{1}{c|}{{0.022}} & $5.718 \times 10^{-3}$ & \multicolumn{1}{|c|}{{0.05}} \\\hline

\end{tabular}
\end{table*}

\subsection{Performance Comparison in Experiments}
Fig. \ref{fig:e_load} plots the instantaneous communication load over time for both the Traditional Wireless Framework and our GSC framework, recorded during experiments. Similar to the simulations, the FS algorithm (second subfigure) perfectly segments the task into different phases as expected, and optimises the communication load in each time slot by only transmitting the GSC information. Building upon this, the PPDQN algorithm (bottom two subfigures) trained in the simulations can further capture the temporal features of the DT and discard unnecessary reconstruction messages to further minimise the communication load. 

Fig. \ref{fig:e_angle} and Fig. \ref{fig:e_vel} plot the normalised errors in joint angle and joint velocity versus time under baselines and our proposed GSC framework in experiments.
Firstly, it can be observed that noticeable reconstruction errors exist all the time in the experiments, which is different from simulations. The reason is that the simulator relies on ideal kinematics and dynamics models of the robot arm, but cannot capture external factors such as joint friction, that might impact reconstruction accuracy in practice.
Secondly, we observe that the error performance of the GSC framework under strict reconstruction error constraints almost completely coincides with that of the Traditional Wireless Framework, which shown the effectiveness of our proposed framework for DT reconstruction. 

To extend the applicability of the proposed GSC framework, we also analysed its performance under non-ideal downlink channel conditions. We observe that the performance of the proposed GSC framework degrades to that of the GSC Framework with Feature Selection but without Temporal Selection when the ideal downlink wireless channel is replaced by a non-ideal one. This degradation is because the PPDQN algorithm  tends to transmit all packets  without any selection under such conditions. Initially, the agent incorrectly concludes that reconstruction message transmission will not improve the reconstruction error due to ACK feedback failure, and then drops most of the messages to maximize the reward. However, when the reconstruction error constraints are violated due to the reduced transmissions, the agent resumes transmissions. At this time, the agent only learns that transmission is useful for reducing the reconstruction errors, but it fails to evaluate the temporal importance of messages. Ultimately, it adopts an overly conservative policy to transmit every packet.

\subsection{Overall Performance Comparison}
Finally, Table \ref{tab:performance} presents a comprehensive comparison of the cumulative communication load, average angle error, and average velocity error for the baselines and our proposed GSC framework in both simulation and practical experiments, which are recorded after the robot arm completes the robotic task once \footnote{For safety reasons, the maximum robot velocity is manually scaled down during experiments compared to the simulations, which might result in a lower reconstruction error in experiments.}. 
By comparing our framework with the baselines, we obtain the following insights:
1) the FS algorithm can reduce the communication load by at least 36.8\% for the three tasks in simulations without increasing the reconstruction error, this is because its effective design in extracting features that contain the GSC information. Its performance is also stable since the phase identification conditions and transmitted features are accurately preset;
2)  compared to the Traditional Wireless Framework, the PPDQN algorithm further reduces the communication load, which achieves at least a 59.5\% reduction under strict error constraints and up to 80\% under relaxed error constraints. We also notice that slight error violations occur in certain cases, which may result from the introduction of cost into the Q-value function and the agent's overestimation of Q-values during training.
3)  in real-world experiments, the communication load is significantly reduced by 53\% under strict constraints and 74\% under relaxed constraints. However, the PPDQN algorithm performs worse compared to simulations because the models cannot perfectly adapt to the real-world data. Collecting more practical robot interaction data under noisy conditions and extreme cases could potentially improve the model's generalisability in real-world scenarios.

\section{Conclusion}  \label{sec:6}
In this work, we proposed a novel goal-oriented semantic communication (GSC) framework for robot arm reconstruction in the Digital Twin (DT), aiming to minimise the communication load under the reconstruction error constraints. Unlike the traditional reconstruction framework that sends a reconstruction message periodically, we optimise the transmission in both feature domain and time domain. We developed a Feature Selection (FS) algorithm that is used to divide the robotic task into multiple phases and selectively transmit more useful features. Building upon this, we designed a Proportional-Integral-Derivative-based primal-dual Deep Q-Network (PPDQN) algorithm to make message temporal selection.
The effectiveness of our framework was validated through both Pybullet simulations and real-world experiments. The communication between the physical and digital twins was realized through an ideal wired link, assumed error-free and instantaneous in simulations, and implemented via a physical Gigabit Ethernet connection in real-world experiments. A simulated wireless channel was further incorporated at the transmitter side before the wired transmission to realistically emulate practical wireless channel conditions.  For three different robotic tasks, it is shown that our framework can reduce the communication load by at least 59.5\% under strict error constraints and 80\% under relaxed error constraints in simulations. In real-world experiments, our framework still achieved a communication load reduction to 53\% under strict error constraints and 74\% under relaxed error constraints.
\bibliographystyle{IEEEtran}
\bibliography{IEEEabrv,JSAC_Digital_Twin}

\begin{IEEEbiography}[{\includegraphics[width=1in,height=1.25in,clip,keepaspectratio]{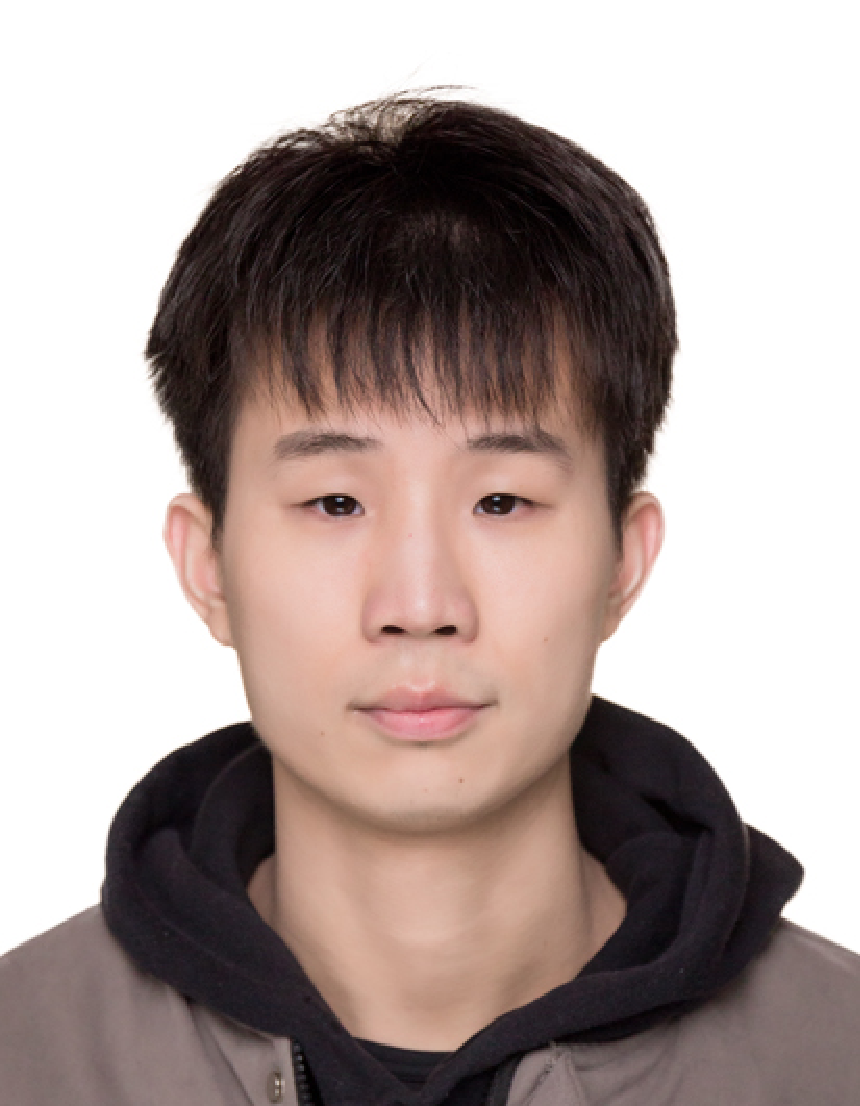}}]
{Shutong Chen} received the B.S. degree from
 Chongqing University, Chongqing, China, in 2022,
 and the M.S. degree from the University College
 London, London, U.K., in 2023. He is currently pursuing the Ph.D. degree with the Center
 for Telecommunications Research, King’s College
 London, London.
 His research interests include semantic communication and digital twin.
\end{IEEEbiography}

\begin{IEEEbiography}[{\includegraphics[width=1in,height=1.25in,clip,keepaspectratio]{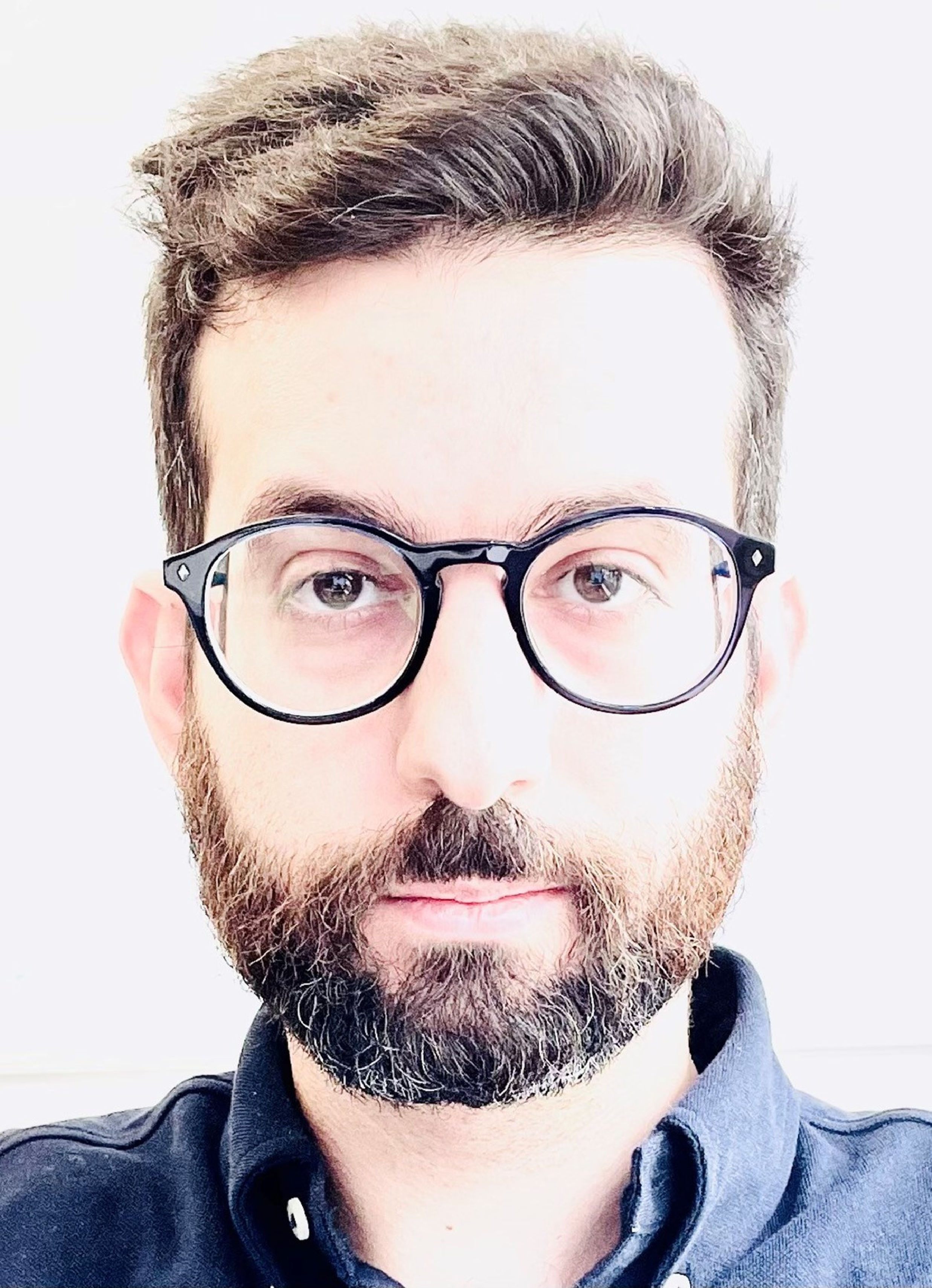}}]
{Emmanouil Spyrakos-Papastavridis} (Member, IEEE) received the B.Sc. degree in electronic and electrical engineering from University College London (UCL), London, U.K., the M.Sc. degree in mechatronics from King’s College London (KCL), London, U.K., and the Ph.D. degree in humanoid robotics from Istituto Italiano di Tecnologia (IIT), Genoa, Italy. He was a Postdoctoral Researcher with the Department of Advanced Robotics, Istituto Italiano di Tecnologia, a visiting researcher at the Dyson School of Design Engineering, Imperial College London, and a Postdoctoral Research Associate with the Department of Engineering, King’s College London, where he was later a Lecturer (Assistant Professor) in Robotics. He is currently a Senior Lecturer (Associate Professor) in Robotics at the Centre for Robotics Research, Department of Engineering, King’s College London. His research interests include articulated-soft robots, nonlinear control, humanoid/bipedal robots, metamorphic robots, exoskeletons, and physical human-robot interaction. Dr. Spyrakos was a recipient of the 2022 Journal of Mechanisms and Robotics Best Paper Award, and a finalist for the Best Interactive Paper Award at Humanoids 2015. He was Principal Investigator of the Engineering and Physical Sciences Research Council (EPSRC) REST project and is currently serving as an Associate Editor for Robotica (Cambridge University Press).
\end{IEEEbiography}

\begin{IEEEbiography}[{\includegraphics[width=1in,height=1.25in,clip,keepaspectratio]{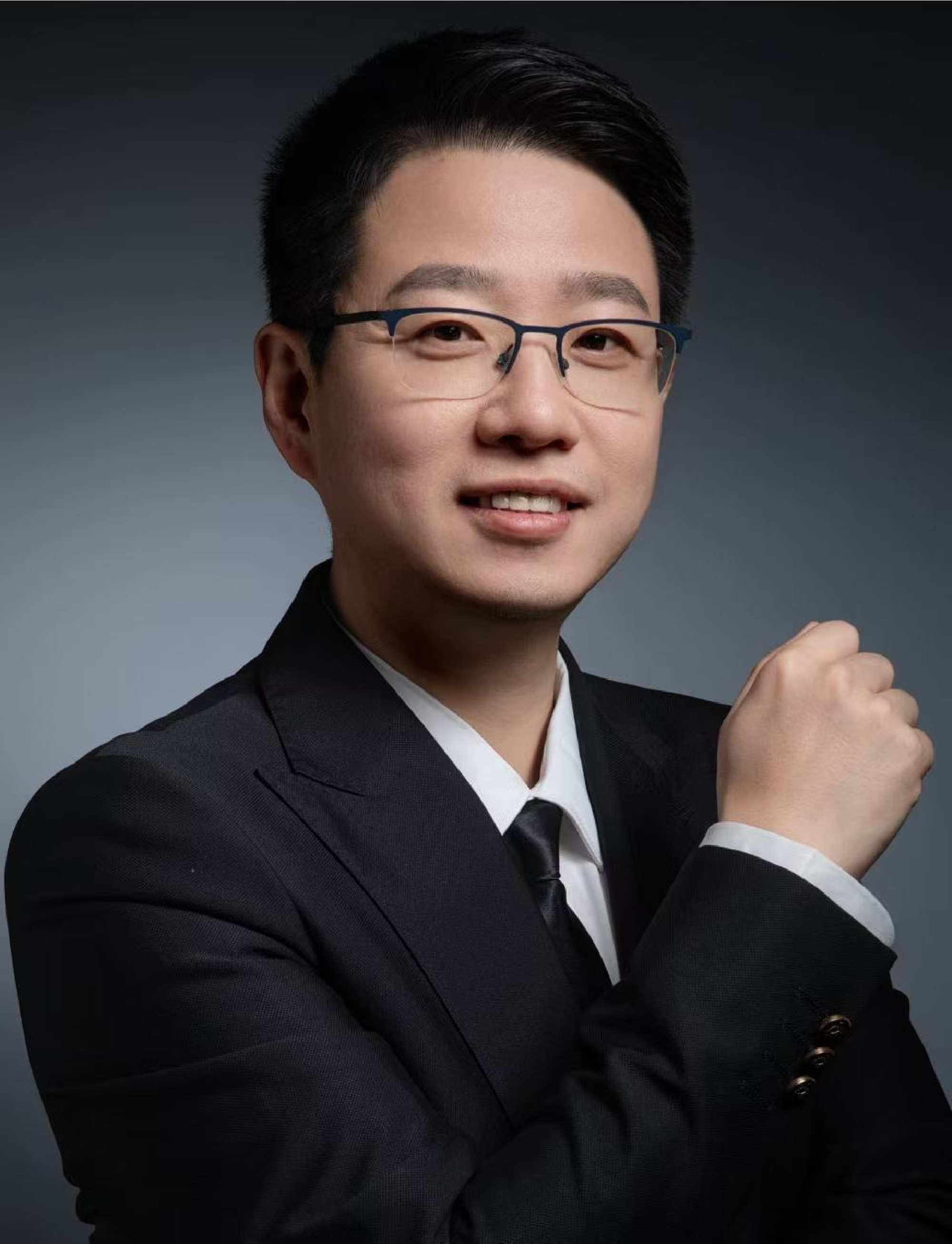}}]
{Yichao Jin} (Senior Member, IEEE) is an industrial automation and wireless communication expert with over 20 years of R\&D experience with international companies such as Toshiba and Samsung. He is Fellow of IET (FIET) and Charted Engineer (CEng), Senior Member of IEEE, and has been previously involved in various of standardization activities such as IETF, IEC and ETSI. He currently works as senior research expert in Zhejiang Lab China focusing on intelligent computing. 
\end{IEEEbiography}

\begin{IEEEbiography}[{\includegraphics[width=1in,height=1.25in,clip,keepaspectratio]{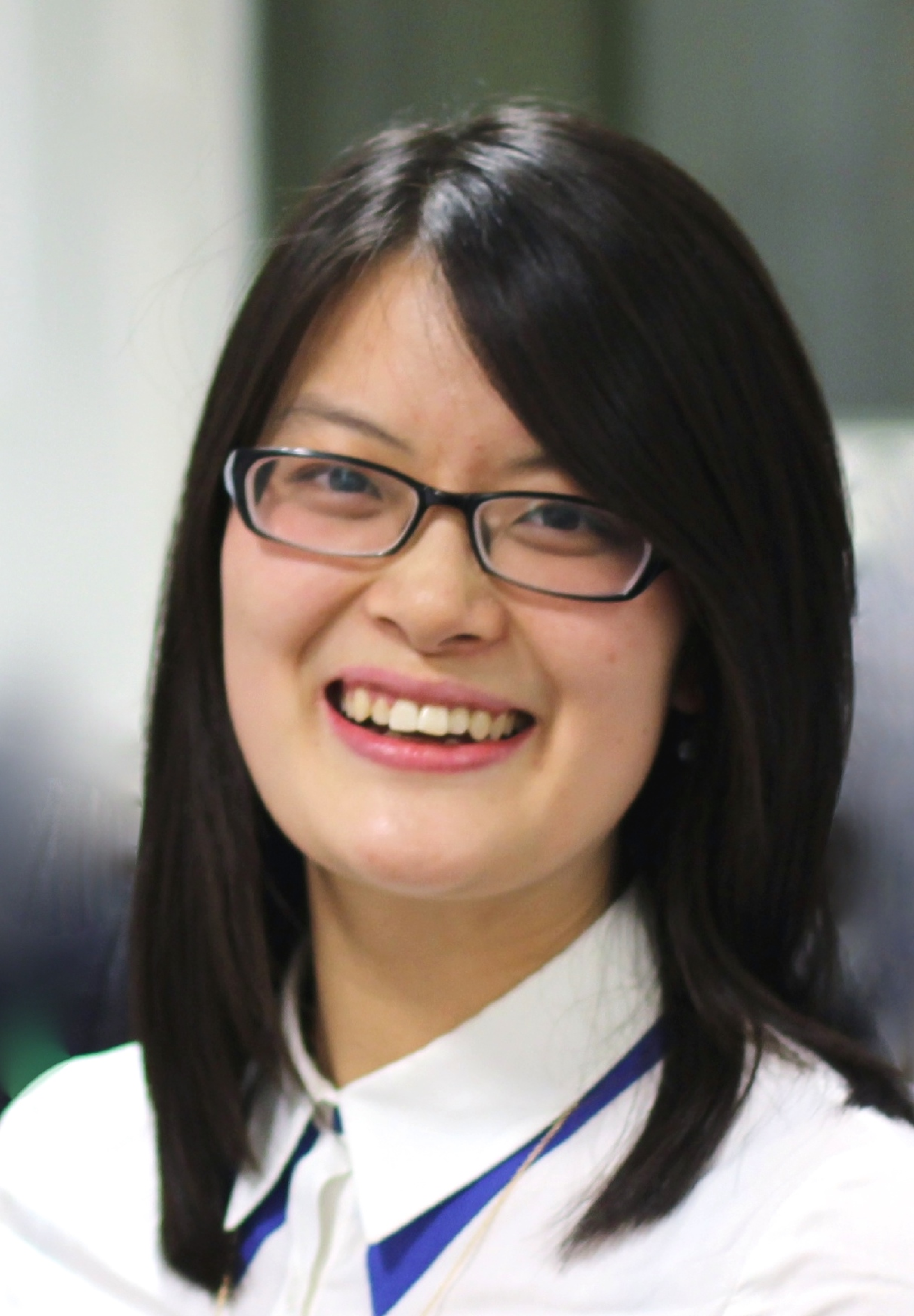}}]
{Yansha Deng} (Senior Member, IEEE) is currently a Reader (Associate Professor) in the Department of Engineering at Kings College London, London, United Kingdom. She received her Ph.D. degree in electrical engineering from the Queen Mary Uni versity of London, U.K., in 2015. From 2015 to 2017, she was a Post-Doctoral Research Fellow with Kings College London, U.K. She has secured more than £2.6 million of research funding as the prin cipal investigator and has received the EPSRC NIA award. She has published 120+ journal papers and 60+ IEEE/ACM conference papers. Her research interests include molecular communication and machine learning for 5G/6G wireless networks. She was a recipient of the Best Paper Awards from ICC 2016 and GLOBECOM 2017 as the first author, and the IEEE Communications Society Best Young Researcher Award for the Europe, Middle East, and Africa Region 2021. She has been the Senior Editor of IEEE Communications Letters since 2020, the Associate Editor of IEEE Transactions on Communications since 2017, the Associate Editor of IEEE Communications Surveys and Tutorials since 2022, the Associate Editor of IEEE Transactions on Machine Learning in Communications and Networking since 2022, the Associate Editor of IEEE Transactions on Molecular, Biological and Multi-scale Communications since 2019, the Associate Editor of IEEE Open Journal of Communications Society since 2019 and the Vertical Area Editor of IEEE Internet of Things Magazine since 2021.
\end{IEEEbiography}

\end{document}